\documentclass[letterpaper,journal]{IEEEtran}
\usepackage{amsmath,amssymb,amsfonts}

\usepackage{placeins}
\usepackage{array}
\usepackage[caption=false,font=normalsize,labelfont=sf,textfont=sf]{subfig}
\usepackage{textcomp}
\usepackage{stfloats}
\usepackage{url}
\usepackage{verbatim}
\usepackage{graphicx}
\usepackage{tikz}
\usepackage{adjustbox}
\usepackage{cite}
\usepackage{xcolor}
\usepackage{tabu}
\usepackage{nomencl}

\usepackage{multirow}
\definecolor{matlabgreen}{RGB}{119,172,48}
\usepackage{subcaption}
\usepackage{tikz}
\definecolor{darkblue}{rgb}{9,0,0}
\definecolor{lightyellow}{rgb}{1, 1, 0.8}
\definecolor{lightblue}{rgb}{0.9,0.9,1}
\usepackage{bm}
\usepackage{threeparttable}
\usepackage{caption}
\usepackage{tabularx}
\definecolor{lightgreen}{rgb}{0.75, 0.95, 0.75} 
\definecolor{lightgray}{rgb}{0.95, 0.95, 0.95}
\usepackage{soul}
\usepackage{booktabs}
\usepackage{listings}
\usepackage{xcolor} 

\definecolor{matlabblue}{RGB}{77,190,238}
\definecolor{matlabblue}{RGB}{77,190,238}
\definecolor{color1}{RGB}{113, 191, 234} 
\definecolor{color2}{RGB}{36, 123, 160} 
\definecolor{matlabpurple}{RGB}{126,47,142}
\definecolor{darkpink}{RGB}{255,19,166}

\lstdefinestyle{mystyle}{
    language=Matlab,
    basicstyle=\ttfamily\footnotesize,
    keywordstyle=\color{blue},
    commentstyle=\color{gray},
    stringstyle=\color{red},
    numbers=left,
    numberstyle=\tiny,
    numbersep=5pt,
    backgroundcolor=\color{white},
    frame=single,
    breaklines=true,
    showstringspaces=false,
    captionpos=b
}
\makenomenclature

\renewcommand{\nompreamble}{The following abbreviations and symbols will be used in the forthcoming equations.}

\usepackage{etoolbox}
\renewcommand\nomgroup[1]{%
  \item[\bfseries
    \ifstrequal{#1}{R}{DNN policy parameters}{%
    \ifstrequal{#1}{S}{RAC policy parameters}{%
    \ifstrequal{#1}{E}{MPD Parameters}{%
    \ifstrequal{#1}{A}{Abbreviations}{%
    \ifstrequal{#1}{Z}{Synthesis of DNN and RAC parameters}{%
    Other Symbols}}}}}%
  ]%
}

\begin{document}

\title{Synthesis of Deep Neural Networks with Safe Robust Adaptive Control for Reliable Operation of Wheeled Mobile Robots}

\author{Mehdi Heydari Shahna, and Jouni Mattila
\thanks{Funding for this research was provided by the Business Finland partnership project ``Future All-Electric Rough Terrain Autonomous Mobile Manipulators'' (Grant No. 2334/31/2022).}
\thanks{All authors are with the Faculty of Engineering and Natural Sciences, Tampere University, 33720 Tampere, Finland. (e-mail: mehdi.heydarishahna@tuni.fi; jouni.mattilia@tuni.fi).}}

\markboth{}%
{Shell \MakeLowercase{\textit{et al.}}: A Sample Article Using IEEEtran.cls for IEEE Journals}


\maketitle

\begin{abstract}
    Deep neural networks (DNNs) can enable precise control while maintaining low computational costs by circumventing the need for dynamic modeling. However, the deployment of such black-box approaches remains challenging for heavy-duty wheeled mobile robots (WMRs), which are subject to strict international standards and prone to faults and disturbances. We designed a hierarchical control policy for heavy-duty WMRs, monitored by two safety layers with differing levels of authority. To this end, a DNN policy was trained and deployed as the primary control strategy, providing high-precision performance under nominal operating conditions. When external disturbances arise and reach a level of intensity such that the system performance falls below a predefined threshold, a low-level safety layer intervenes by deactivating the primary control policy and activating a model-free robust adaptive control (RAC) policy. This transition enables the system to continue operating while ensuring stability by effectively managing the inherent trade-off between system robustness and responsiveness. Regardless of the control policy in use, a high-level safety layer continuously monitors system performance during operation. It initiates a shutdown only when disturbances become sufficiently severe such that compensation is no longer viable and continued operation would jeopardize the system or its environment. The proposed synthesis of DNN and RAC policy guarantees uniform exponential stability of the entire WMR system while adhering to safety standards to some extent. The effectiveness of the proposed approach was further validated through real-time experiments using a 6,000 kg WMR.
\end{abstract}
\def\abstractname{Note to Practitioners}
\begin{abstract}
The increasing demand to reduce labor costs and automate complex, repetitive tasks in industrial and construction applications is driving the transition from heavy-duty machinery to large-scale robots. As a result, heavy-duty wheeled mobile robots (WMRs) equipped with complex actuation mechanisms are expected to operate in harsh environments and perform high-power dynamic tasks. Deep neural networks (DNNs) can enable precise control with low computational costs at deployment by eliminating the need for complex modeling in such applications. However, a major challenge arises from the inherent difficulty of ensuring interpretability and analyzing the stability of black-box, non-deterministic learning-based control policies. This challenge becomes particularly pronounced in heavy-duty WMRs, which are susceptible to faults and external disturbances while needing to adhere to strict safety and reliability standards such as ISO 3691, ISO 13849, and IEC 61508. Motivated by Section 10 of ISO/IEC TR 5469 - Artificial intelligence — Functional safety and AI systems, we address this challenge by designing a novel hierarchical control policy for heavy-duty WMRs equipped with active suspension bogies that feature a serial-parallel structure of actuation chains, monitored by two safety layers with differing levels of authority, without requiring knowledge of the actuator model and wheel–ground interactions. The low-level safety layer deactivates the initial high-accuracy DNN-based control policy when the WMR encounters severe external disturbances, activating a supervisory robust adaptive control (RAC) strategy as an alternative. This compensates for the effects of the disturbances and enables continued safe operation by enhancing system robustness, albeit at the cost of reduced accuracy for the remainder of the operational period. A high-level safety layer continuously monitors system performance under disturbances and allows the operation to continue if disturbance compensation is deemed feasible; otherwise, to prevent posing a risk to the system or its environment, it triggers a shutdown. Interestingly, synthesizing the proposed RAC with the high-accuracy DNN control policy improves the interpretability, robustness, and stability of the WMR within the proposed hierarchical control architecture. This approach could pave the way for broader use of learning algorithms in heavy-duty robotic systems while still meeting stringent international standards. The approach is further verified step-by-step through experimental evaluation of the standalone DNN-based policy, the standalone RAC policy, and their synthesis on a 4,836 kg WMR equipped with a five-stage actuation mechanism. The comparative results demonstrate the effectiveness of the proposed control framework.
\end{abstract}


\nomenclature[A]{\(\textbf{DNN}\)}{Deep neural network}
\nomenclature[A]{\(\textbf{WMR}\)}{Wheeled mobile robot}
\nomenclature[A]{\(\textbf{BLF}\)}{Barrier Lyapunov function}
\nomenclature[A]{\(\textbf{RAC}\)}{Robust adaptive control}
\nomenclature[A]{\(\textbf{MPD}\)}{Multi-purpose deployer}
\nomenclature[A]{\(\textbf{PMSM}\)}{Permanent magnet synchronous motor}
\nomenclature[A]{\(\textbf{LM}\)}{Levenberg–Marquardt}
\nomenclature[A]{\(\textbf{MSE}\)}{Mean squared error}
\nomenclature[A]{\(\textbf{Tanh}\)}{Hyperbolic tangent}
\nomenclature[A]{\(\textbf{rpm}\)}{Revolutions per minute}
\nomenclature[A]{\(\textbf{PPC}\)}{Prescribed performance control}
\nomenclature[A]{\(\textbf{SSWMR}\)}{Skid-steer wheeled mobile robot}
\nomenclature[A]{\(\textbf{rpm}\)}{Revolutions per minute}
\nomenclature[A]{\(\textbf{AI}\)}{Artificial intelligence}

\nomenclature[E]{\(i\)}{Side of the MPD $i=L(left), R(right)$}
\nomenclature[E]{\(D_p\)}{Pump displacement ($\mathrm{m}^3 / \mathrm{rad}$)}
\nomenclature[E]{\(\omega_{p_i}\)}{Angular velocity of PMSM/pump ($\mathrm{m} / \mathrm{rad}$)}
\nomenclature[E]{\(\omega_{h_i}\)}{Angular velocity of hydraulic motor ($\mathrm{m} / \mathrm{rad}$)}
\nomenclature[E]{\(\omega_{w_i}\)}{Angular velocity of hydraulic motor ($\mathrm{m} / \mathrm{rad}$)}
\nomenclature[E]{\(r\)}{Radius of wheel ($\mathrm{m}$)}
\nomenclature[E]{\(Q_i\)}{Hydraulic pump flow ($\mathrm{m}^3 / \mathrm{s}$)}
\nomenclature[E]{\(D_m\)}{Hydraulic motor displacement ($\mathrm{m}^3 / \mathrm{rad}$)}
\nomenclature[E]{\(v_i\)}{Linear velocity of wheel ($\mathrm{m}/ \mathrm{s}$)}
\nomenclature[E]{\(G\)}{Gear reduction}
\nomenclature[E]{\(K_v\)}{Positive constant}
\nomenclature[E]{\(n_{p_i}\)}{Velocity of PMSM/pump (rpm)}
\nomenclature[E]{\(\tau\)}{Positive constant}
\nomenclature[E]{\(F_i\)}{External disturbance and modeling error}
\nomenclature[E]{\(A_i\)}{Positive constant}
\nomenclature[E]{\(v_{{ref}_i}\)}{Desired linear velocity of wheel ($\mathrm{m}/ \mathrm{s}$)}
\nomenclature[E]{\(e_i\)}{Tracking velocity error of wheel ($\mathrm{m}/ \mathrm{s}$)}
\nomenclature[R]{\(L\)}{Total number of hidden layers}
\nomenclature[R]{\(\ell\)}{Index the current hidden layer}
\nomenclature[R]{\(W^{(\ell)}\)}{The weight matrix for layer $\ell$}
\nomenclature[R]{\(n_{\ell}\)}{Number of neurons in layer $\ell$}
\nomenclature[R]{\(b^{(\ell)}\)}{Bias vector associated with layer $\ell$}
\nomenclature[R]{\(z^{(\ell)}\)}{Pre-activation vector of layer $\ell$}
\nomenclature[R]{\(a^{(\ell)}\)}{Activation vector of layer $\ell$}
\nomenclature[R]{\(\varphi^{(\ell)}\)}{Activation function applied at layer $\ell$}
\nomenclature[R]{\(\mathcal{V}_i\)}{Batch input matrix for the $i$th side of the MPD}
\nomenclature[R]{\(P\)}{Number of training samples (or batch size)}
\nomenclature[R]{\(\bm{1}\)}{A column vector of ones}
\nomenclature[R]{\(E_{\text{MSE}}\)}{MSE loss function}
\nomenclature[R]{\(N_{p_i}\)}{Matrix of predicted outputs}
\nomenclature[R]{\(T_{p_i}\)}{Matrix of target (true) outputs}
\nomenclature[R]{\(t_{p_i}\)}{Target output}
\nomenclature[R]{\(\theta\)}{Set of all trainable parameters}
\nomenclature[R]{\(F\)}{Frobenius norm}
\nomenclature[R]{\(\xi_{p_i}^{(p)}\)}{Prediction error for the $p$-th training sample}
\nomenclature[R]{\(N_w\)}{Total number of trainable parameters}
\nomenclature[R]{\(w\)}{Flattened parameter vector}
\nomenclature[R]{\(\xi_{p_i}\)}{Stacked error vector over all training samples}
\nomenclature[R]{\(m\)}{Dimension of the network's output layer}
\nomenclature[R]{\(J\)}{Jacobian matrix of the stacked error vector}
\nomenclature[R]{\(s_k\)}{$k$-th scalar component of the stacked error vector}
\nomenclature[R]{\(j\)}{Index the trainable parameters in the neural network}
\nomenclature[R]{\(k\)}{Index the scalar of the stacked error vector}
\nomenclature[R]{\(H\)}{Hessian matrix of the MSE loss function}
\nomenclature[R]{\(\mu\)}{Damping parameter in the LM optimization}
\nomenclature[R]{\(\eta\)}{Step size in the gradient descent interpretation}
\nomenclature[R]{\(I\)}{Identity matrix}
\nomenclature[R]{\(b\)}{Iteration index in the optimization process}
\nomenclature[R]{\(w_{cand}\)}{Candidate new weights}
\nomenclature[R]{\(\beta\)}{Damping adjustment factor in the LM algorithm}
\nomenclature[R]{\(\beta\)}{Damping adjustment factor in the LM algorithm}
\nomenclature[R]{\(\zeta\)}{PPC bound on the tracking error $e_i$}
\nomenclature[R]{\(\zeta^{shoot}\)}{Allowable overshoot in the PPC framework}
\nomenclature[R]{\(\zeta^{bound}\)}{Final steady-state error bound in the PPC framework}
\nomenclature[R]{\(\zeta^{*}\)}{Convergence rate bound in the PPC framework}

\nomenclature[S]{\(o\)}{PPC bound on the tracking error $e_i$}
\nomenclature[S]{\(o^{shoot}\)}{Allowable overshoot in the PPC framework}
\nomenclature[S]{\(o^{bound}\)}{Final steady-state error bound in the PPC framework}
\nomenclature[S]{\(o^{*}\)}{Convergence rate bound in the PPC framework}
\nomenclature[S]{\(\hat{\theta}_i\)}{Adaptive law parameter for the $i$-th side}
\nomenclature[S]{\(\tilde{\theta}_i\)}{Adaptive law error for the $i$-th side}
\nomenclature[S]{\({\theta}^*_i\)}{Unknown positive constant}
\nomenclature[S]{\(\delta_i\)}{Positive constant}
\nomenclature[S]{\(\gamma_i\)}{Positive constant}
\nomenclature[S]{\(V_i\)}{Positive quadratic function for the $i$-th side}
\nomenclature[S]{\(f^*_i\)}{Unknown positive bound}
\nomenclature[S]{\(\epsilon_i\)}{Arbitrary positive constant}
\nomenclature[S]{\(\mu_i\)}{Unknown positive constant}
\nomenclature[S]{\(\ell_i\)}{Unknown positive bound}

\nomenclature[Z]{\(u_{s_i}\)}{RAC policy command for $n_{p_i}$}
\nomenclature[Z]{\(u_{{DNN}_i}\)}{DNN policy command for $n_{p_i}$}
\nomenclature[Z]{\(\alpha_1\)}{Latched-OFF logic}
\nomenclature[Z]{\(\alpha_2\)}{Latched-ON logic}
\nomenclature[Z]{\(\zeta\)}{Performance metric in low-level safety layer}
\nomenclature[Z]{\(o\)}{Performance metric in high-level safety layer}

\printnomenclature

\section{Introduction}
\subsection{Background and Context}
\label{background}

\IEEEPARstart{T}{he} concept of robotization traces back to ancient and Renaissance-era fascination with self-operating machines, eventually evolving into the modern vision of autonomous robots \cite{herman1998renaissance}. Coined in 1920, the term \textit{“robot”} reflected the aspiration to create artificial laborers. By the mid-20th century, industrial robots began performing repetitive tasks in factories, driven by the goals of increased productivity, precision, and safety \cite{romanenko2022robot}. Today, robots operate continuously, excel at hazardous or monotonous tasks, reduce error rates, and enable humans to focus on higher-level functions \cite{adams2025human}. Robotization has also transformed wheeled construction machinery by converting traditional human-operated equipment into autonomous wheeled mobile robots (WMRs) \cite{klancar2017wheeled, machado2021autonomous}.
Heavy-duty construction machines perform high-power tasks in hazardous, unstructured environments like tunnels, mines, and construction sites, where unstable terrain, limited visibility, and physical dangers pose risks to both equipment and operators. Certain areas, such as deep pits or confined tunnel interiors, are particularly unsafe or inaccessible to humans due to spatial and environmental hazards \cite{chi2017avoiding}. These factors underscore the need for intelligent autonomous or semi-autonomous control systems that can maintain operational efficiency while ensuring safety in high-risk and hard-to-reach conditions \cite{leung2023automation}.
Skid-steer wheeled mobile robots (SSWMRs) are widely adopted in off-road and construction environments \cite{zuo2022visual} since their mechanical configuration eliminates complex steering linkages and allows for in-place pivot turns, essentially achieving a zero turning radius \cite{plonski2016environment}.
This design provides SSWMRs with extraordinary agility in confined spaces and tight job sites while offering exceptional maneuverability and multi-functional versatility on difficult terrain, along with cost-efficient, low-maintenance operation \cite{zhao2018adaptive}. In addition, most heavy-duty SSWMRs employ in-wheel hydraulic motors due to their superior torque density and mechanical robustness in withstanding harsh operating environments \cite{li2024energy}. Hydraulic motors are particularly well-suited for mobile applications requiring high force output at low speeds, such as traction and navigation in off-road or uneven terrain \cite{shahna2025robudfdsfst}. To power these hydraulic motors, modern electro-hydraulic architectures increasingly employ permanent
magnet synchronous motors (PMSMs) due to their high power density, precise controllability, and, above all, superior efficiency compared to conventional machines \cite{jin2023artificial, shahasaasca2024asdacvrobustness}. Their ability to deliver smooth torque at low speeds is particularly beneficial for modulating hydraulic pressure and flow under varying load conditions.

\subsection{Related Work, Research Gap, and Motivations of the Study}
\label{Motivations}

Model-based control of SSWMRs has become a prominent focus in both academic and industrial research due to its critical role in autonomous navigation \cite{liao2017performance, rizk2023model}.
In control theory, the term model-based refers to the extent to which knowledge of a system’s dynamics is utilized in controller design. When an accurate model is available, model-based controllers can offer high performance and precision \cite{zhang2021model, precup2021data}. Hence, the authors in \cite{li2015trajectory} proposed a model predictive control scheme incorporating neural-dynamic optimization to achieve accurate trajectory tracking of a two-wheel SSWMR. Similarly, a linear quadratic regulator in \cite{elshazly2014skid} and a virtual force control in \cite{mazur2015virtual} were proposed by incorporating the dynamic model of four-wheel-driven configurations in SSWRMs, where the wheels on each side were mechanically coupled. As is evident, many state-of-the-art model-based control strategies rely heavily on the availability of an accurate mathematical model of the system—typically in the form of a transfer function or state-space representation. 
Unlike lightweight robots, heavy-duty WMRs often incorporate multi-stage actuation mechanisms to transmit power from the energy source to the ground-contacting wheels, based on control commands, to generate motion \cite{shahna2025lidar, shahna2025robudfdsfst}. Each stage of an actuation chain contributes distinct dynamic properties that interact nonlinearly and often with strong kinematic coupling.
These interdependencies result in complex input-output relationships, where disturbances or delays in one stage propagate through the system and degrade overall performance. This makes parameter-rich modeling of heavy-duty WMRs analytically complex and computationally intensive to develop and validate. As a result, many model-based control approaches omit the effects of wheel–ground interactions \cite{liao2018model} and other difficult-to-characterize system behaviors, instead aiming to enhance system robustness through adaptive control techniques, such as those presented in \cite{liao2017performance, xiong2022path, mohammadpour2011robust}. Although these techniques increase robustness, they can degrade tracking accuracy due to a fundamental trade-off between disturbance rejection and system responsiveness, wherein improving one often comes at the expense of the other \cite{shahasaasca2024asdacvrobustness, chen2025cascade}.

To address such challenges, deep neural networks (DNNs) can aid in capturing the dynamic behavior of a robot while reducing computational effort due to their powerful approximation and generalization capabilities \cite{li2021dnn}. Unlike traditional physics-based models, which require detailed knowledge of system parameters, DNNs can learn complex, nonlinear relationships directly from sensory data, including hard-to-model phenomena such as friction, wheel–ground interaction, and actuator delay \cite{badgujar2023deep}. Once trained, a DNN can replace or augment computationally expensive dynamic equations with a lightweight inference process, significantly reducing the real-time computational load. This makes DNNs especially valuable for high-dimensional systems such as heavy-duty SSWMRs equipped with active suspension bogies featuring a multi-stage structure, where traditional model-based control may become analytically intractable or computationally prohibitive \cite{costello2023dnn, kuo2022deploy}. Despite the high accuracy achieved by DNNs in learning complex dynamical models within their trained domains, control strategies based on DNNs inherently lack robustness when exposed to unknown forces or operating conditions outside the training distribution \cite{chaudhury2021robustness, yu2020rein}. This limitation arises primarily from the data-driven nature of DNNs, which generalize reliably only within the statistical boundaries of their training data. Moreover, collecting data that captures all possible operating conditions a robot might encounter is often impractical in terms of time and computational effort—especially since some conditions may never occur in practice, may occur only rarely, or may be overlooked during the data collection process \cite{amir2024verifying}. In addition, unlike model-based controllers, learning-based control policies have a black-box nature, making their interpretability limited and formal stability analysis highly challenging \cite{csahin2025unlocking, agarwal2024hybrid}. Consequently, guaranteeing reliable performance under uncertain or safety-critical conditions remains a significant challenge \cite{shahna2025lidar}. This inability to ensure performance and operational reliability is unacceptable for heavy-duty machinery, which must comply with stringent international safety and functional standards such as \textcolor{black}{ISO 3691 \cite{iso1977iso}}, \textcolor{black}{ISO 13849 \cite{main2014control}}, and \textcolor{black}{IEC 61508 \cite{bell2006introduction}} to prevent harm to both the environment and personnel. This concern is particularly critical given that such robotic systems are expected to operate in harsh, unstructured environments and execute high-power, dynamic tasks, making them inherently susceptible to external disturbances, faults, and failure modes.

In response to the aforementioned challenges, this study aims to address a fundamental question: How can learning-based control policies—designed to bypass analytically intensive modeling while achieving high performance under nominal conditions—be effectively deployed on heavy-duty WMRs equipped with complex actuation structures, in a manner that guarantees safety and system stability in accordance with relevant international safety standards? In formal control theory, safety bounds for dynamic and hybrid systems are typically established using techniques such as barrier Lyapunov functions (BLFs) \cite{wang2023concurrent, liang2023adaptive}. These methods ensure that the system remains within a predefined safe operating region. However, the resulting sets are often conservative, as they do not actively drive the system back toward safer states once the boundary is approached \cite{liu2019barrier}.

\subsection{Contributions and Structure of the Paper}
\label{contriii}
Motivated by Section 10 of ISO/IEC TR 5469 - Artificial intelligence (AI) — Functional safety and AI systems \cite{iso5469}, supervisory barrier functions can serve as monitoring mechanisms to detect potentially unsafe behavior caused by AI technologies, whether due to internal faults or external disturbances. Upon detection, the system can initiate conventional stabilizing controllers to maintain safety. Motivated by this standard, we propose a hierarchical control policy for heavy-duty WMRs, monitored by two safety layers with differing levels of authority. To this end, a high-performance DNN control policy is trained using data collected from onboard sensors, capturing the relationship between actual motion and control commands. These data are generated through simple open-loop controller, in which control command inputs are gradually increased to their maximum values in a safe environment under human supervision to record the corresponding system response. The resulting trained DNN control is then deployed as the initial control policy during normal operation, providing high-precision performance. If external disturbances occur and become sufficiently intense, such that the control performance fails to meet a prescribed threshold, the low-level safety layer deactivates the DNN-based control policy and activates a model-free robust adaptive control (RAC) strategy as an alternative, ensuring continued safe operation. By managing the fundamental trade-off between system robustness and responsiveness, this transition compensates for the effects of external disturbances and enables continued safe operation by enhancing robustness, albeit at the cost of reduced system responsiveness for the remainder of the operational period. Regardless of the control policy in use, a logarithmic BLF is employed as a high-level safety layer that continuously monitors system performance during operation. A shutdown is triggered only when disturbances become sufficiently severe such that compensation is no longer feasible and continued operation would pose a risk to the system or its environment. The key findings of this research are summarized as follows:

1) High-precision tracking performance is achieved through the synthesis of a DNN and a safe RAC, while simultaneously ensuring robustness against external disturbances, in accordance with ISO/IEC TR 5469. This eliminates the need for explicit knowledge of the complex actuation mechanisms inherent in heavy-duty SSWRMs

2) Demanding control tracking error thresholds, such as overshoot and steady-state error, are adjustable within two different levels of safety layers. This adaptability allows the system to accommodate the specific performance requirements of the target application.

3) The proposed control framework guarantees the uniform exponential stability of the entire SSWRM system, even in the presence of the black-box learning-based model and external disturbances.

The rest of the paper is structured as follows.
Section \ref{section:energy_conversion} details the system modeling and problem formulation. Section \ref{section:trajectory_optimization} describes the design of a DNN-based control policy specifically for SSWRMs operating under nominal conditions, leveraging previously recorded sensor data. To enhance the system's robustness against external disturbances, Section \ref{section:state_space_EMLA} outlines the proposed model-free RAC, which includes an adaptive law for each side of the robot. Both these control policies are meticulously developed within a two-layer safety architecture in Section \ref{aasafadfgadA}, with each layer operating at a distinct level of authority. In Section \ref{expmsdnaf}, the approach is further verified through a step-by-step experimental evaluation of the standalone DNN-based policy, the standalone RAC policy, and their synthesis on a 6,000 kg WMR equipped with a five-stage actuation mechanism. The comparative results demonstrate the feasibility of all proposed approaches in real-world scenarios. Section \ref{section:conclusion} provides a concise summary of the study's key findings and presents the overarching conclusions derived from this research. Detailed MATLAB coding instructions are provided in the \textit{Appendices} to facilitate replication and deeper understanding of the implemented methodologies.

\section{System Modeling and Problem Formulation}
\label{section:energy_conversion}
\begin{figure}[h] 
  \centering
\scalebox{1}
    {\includegraphics[trim={0cm 0.0cm 0.0cm
    0cm},clip,width=\columnwidth]{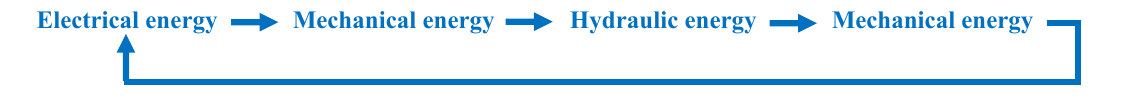}}
  \caption{A multi-stage actuation system for each side of the PMSM-powered SSWMR.}
  \label{fig111sadsad521f}
\end{figure}

Following Section \ref{background}, consider a PMSM-powered heavy-duty SSWMR equipped with two identical multi-stage actuation mechanisms—one on each side—that convert electrical and hydraulic energy into mechanical motion, as illustrated in Fig. \ref{fig111sadsad521f}.

\begin{figure} [t]
    \centering
    \scalebox{1}{\includegraphics[trim={0cm 0.0cm 0.0cm 0cm},clip,width=\columnwidth]{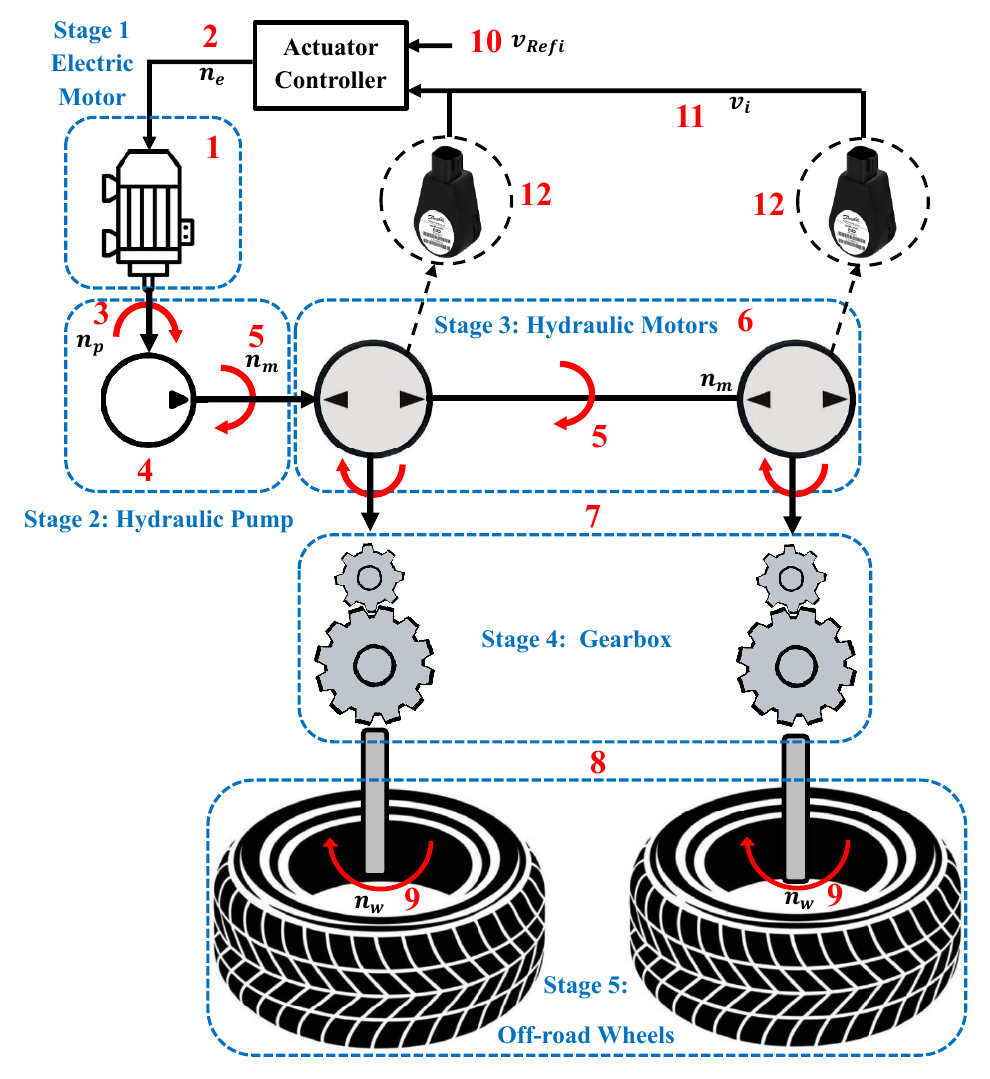}}
  \caption{Identical active suspension bogies on each side of the SSWMR. \textcolor{red}{1}	Electric motor drives the system; \textcolor{red}{2} Control command (rpm) for electric motor; \textcolor{red}{3} The generated velocity (rpm) of the hydraulic pump coupled with the electric motor; \textcolor{red}{4}	Hydraulic pump coupled with the electric motor to power the hydraulic system; \textcolor{red}{5} The generated velocity (rpm) of hydraulic motors; \textcolor{red}{6} Serial hydraulic system (two identical hydraulic motors); \textcolor{red}{7} Gearboxes with identical gear ratios; \textcolor{red}{8} Two identical wheels; \textcolor{red}{9} The generated wheel velocity (rpm); \textcolor{red}{10} Reference wheel velocity ($m/s$); \textcolor{red}{11} Measured linear velocity ($m/s$); \textcolor{red}{12} Speed sensors}
  \label{aFfa} 
\end{figure}

Such applications often feature two identical active suspension bogies on each side, featuring a serial-parallel actuation structure integrated with linear actuators. Each side is driven by the PMSM, which functions as the primary force generator for the entire mechanism and can serve as the control input to the SSWMR, depending on the PMSM configuration, it can be controlled via its signals—primarily its rotational speed (rpm) or output torque (Nm). Hence, the PMSM drives a hydraulic pump that supplies power to two serially connected hydraulic motors, which are mechanically coupled and typically operate at the same angular velocity. These hydraulic motors deliver torque through a gearbox to drive the wheels on the corresponding side of the robot. A velocity sensor, mounted on either one of the hydraulic motors or directly on the wheels, measures angular velocity and provides real-time feedback for control purposes, as illustrated in Fig.~\ref{aFfa}.
Assume the primary control objective is to design control laws that generate rpm commands for the PMSMs such that the resulting wheel velocities on each side accurately track the desired velocity profiles, thereby ensuring stable and precise motion control of the platform. Let $i=R, L$ denote the left and right sides of the SSWMR, respectively. Since, the pump is directly coupled to the shaft of the PMSM (no gearbox, no belt drive), the pump flow rate is often proportional to the angular electric motor and hydraulic pump speeds $\omega_{e_i} = \omega_{p_i}:\mathbb{R}^{+} \rightarrow \mathbb{R}$, as

\begin{equation}
\begin{aligned}
\small
\label{1}
Q_i=D_p \cdot \omega_{p_i}
\end{aligned}
\end{equation}
where $D_p \in \mathbb{R}$ is the pump displacement ( $\mathrm{m}^3 / \mathrm{rad}$ ). Assuming incompressible fluid and negligible leakage, the flow through the series hydraulic motors equals the pump flow:

\begin{equation}
\begin{aligned}
\small
\label{1}
Q_i=D_m \cdot \omega_{h_i} \Rightarrow \omega_{h_i}=\frac{D_p}{D_m} \cdot \omega_{p_i}
\end{aligned}
\end{equation}
where $D_m \in \mathbb{R}$ is the motor displacement ( $\mathrm{m}^3 / \mathrm{rad}$ ) and $\omega_{h_i}:\mathbb{R}^{+} \rightarrow \mathbb{R}$ is the angular velocity of both hydraulic motors. Since the motors are in series and mechanically coupled, they rotate at the same speed. Then, the hydraulic motors drive the wheels through a gear reduction $G \in \mathbb{R}^+$. Hence, the angular velocity of each side is obtained as
\begin{equation}
\begin{aligned}
\small
\label{3}
&\omega_{w_i}=\frac{1}{G} \cdot \omega_{h_i}=\frac{D_p}{G D_m} \cdot \omega_{p_i}
\end{aligned}
\end{equation}
Thus, the linear velocity of the wheel is
\begin{equation}
\begin{aligned}
\small
\label{4}
v_i=r \cdot \omega_{w_i}=\frac{r D_p}{G D_m} \cdot \omega_{p_i}
\end{aligned}
\end{equation}
where $r \in \mathbb{R}^+$ is the wheel radius (assume wheels have the same radius). Since the velocity of the pump/PMSM in rpm, denoted by $n_{p_{i}} = n_{e_{i}}:\mathbb{R}^{+} \rightarrow \mathbb{R}$, serves as the control input signal, substituting $\omega_{p_i}=\frac{2 \pi}{60} n_{p_i}$ yields
\begin{equation}
\begin{aligned}
\small
\label{5}
v_i=\frac{r D_p}{G D_m} \cdot \frac{2 \pi}{60} \cdot n_{p_i}=K_v \cdot n_{p_i}
\end{aligned}
\end{equation}
where $K_v=\frac{2 \pi r D_p}{60 G D_m} \in \mathbb{R}^+$ is a positive constant. Eq. \eqref{5} implicitly assumes that any change in the PMSM speed $n_{p_i}$ results in an immediate change in the wheel's linear velocity $v_i:\mathbb{R}^{+} \rightarrow \mathbb{R}$, without accounting for delays or dynamic behavior. In practice, however, systems involving hydraulic and mechanical components exhibit considerable inertia and time delays, which must be considered for realistic modeling and effective control design. The simplest way to add dynamics is to use a first-order differential equation, as
\begin{equation}
\begin{aligned}
\small
\label{6}
\tau \frac{d v_i}{d t}+v_i=K_v \cdot n_{p_i}
\end{aligned}
\end{equation}

This dynamic equation is mathematically equivalent to a first-order low-pass filter, which smooths out fast changes. Eq. \eqref{6} indicates that the rate of change of velocity $\frac{d v_i}{d t}:\mathbb{R}^{+} \rightarrow \mathbb{R}$ affects how fast $v_i$ responds. $\tau \in \mathbb{R}^+$ is a positive time constant, determining how ``slow" or ``fast" the response is. $\tau>0$ means that the system output $v_i$ lags behind the input.

\textit{Assumption II.1.} Setting $\tau$ to zero is not physically realistic, as all real-world systems exhibit inertia and delays that prevent instantaneous response.

Eq. \eqref{6} still represents a highly simplified model and neglects several critical factors that significantly affect the system dynamics. It disregards system uncertainties, actuation inefficiencies, and external disturbances such as wheel–ground interactions and slippage. Based on \cite{shahna2025robudfdsfst}, we assume an unknown nonlinear function $F_i(v_i, t):\mathbb{R}^{+} \times \mathbb{R} \rightarrow \mathbb{R}$, corresponding the external disturbances; thus,

\begin{equation}
\begin{aligned}
\small
\label{7}
\tau \frac{d v_i}{d t}+v_i=K_v \cdot n_{p_i} + F_i(v_i,t)
\end{aligned}
\end{equation}
As we aim to design a model-free control framework, we assume all functions and coefficients $F_i$, $\tau$, and $K_v$ are bounded and unknown (see Assumption II.3). Thus, we can express Eq. \eqref{7}, as
\begin{equation}
\begin{aligned}
\small
\label{8}
\dot{v}_i(t)=A_i n_{p_i}(t)+\tau^{-1} (F_i - v_i)
\end{aligned}
\end{equation}
where $A_i = \frac{K_v}{\tau} \in \mathbb{R}^+$. Let $v_{{ref}_i}:\mathbb{R}^{+} \rightarrow \mathbb{R}$ denote the reference velocity for each side of the SSWMR, generated by the motion planning section in accordance with the application's objectives.
If we define the velocity tracking error of each side $e_i = v_i -v_{{ref}_i} :\mathbb{R}^{+} \times \mathbb{R} \times \mathbb{R} \rightarrow \mathbb{R}$, based on Eq. \eqref{8}, we have
\begin{equation}
\begin{aligned}
\small
\label{9}
& \dot{e}_{_{i}} = A_i n_{p_i} +\tau^{-1} (F_i - v_i)- \dot{v}_{{ref}_i}\\
\end{aligned}
\end{equation}

\textit{Assumption II.2.}
The desired acceleration of each side of the SSWMR, denoted by $\dot{v}_{{ref}_i} $ is bounded. Moreover, the control input gain $A_i$ is known to be positive, ensuring that the direction of the control input is consistent with the motion of the SSWMR.

\textit{Assumption II.3.} We note that $\tau$ and $v_i$ are inherently bounded due to their physical nature, and $\dot{v}_{{ref}_i}$ is assumed to be bounded based on \textit{Assumption II.2}. Consistent with established results in the robust control literature (e.g., \cite{zhang2019low, huang2019practical}), we further assume that the functional gain $A_i$ and the dynamic uncertainty term $F_i$ are each bounded above by unknown but finite positive constants. This implies that for all admissible values of $v_i$ and for all $t$, neither $A_i$ nor $F_i$ can grow unbounded. Allowing unbounded uncertainties would require arbitrarily large, potentially infinite control inputs to guarantee robustness and stability, which is an impractical and unsafe demand in real-world systems. Such excessive control efforts could lead to actuator saturation, mechanical failure, or trigger built-in emergency shutdown protocols. To address these practical constraints, we later introduce safety-constrained control design strategies that explicitly account for and respect these physical and operational limitations.

\textit{Remark II.1.} In this section, we presented the modeling parameters of a heavy PMSM-powered SSWMR as a practical example to illustrate the computational challenges associated with accurately identifying all stages of system parameters. Despite these complexities, our objective is to develop a model-free control framework that operates independently of any detailed system modeling and is not limited to PMSM-powered SSWMR applications. This approach treats the system as a black box—accepting velocity or torque signals from the primary force generator as control inputs and receiving wheel motion feedback as the system output—thereby eliminating the need for explicit knowledge of the internal dynamics.

\section{Safe DNN-based Control Policy}
\label{section:trajectory_optimization} 
Given that the dynamics of each side of the actuation mechanism of the SSWMR involve a single input and a single output, we adopt an independent feed-forward DNN policy for each side. These policies are trained using data acquired from onboard sensors, which capture the nonlinear characteristics of the actuation mechanism—that is, the relationship between the control inputs ($n_{p_i}$) and the resulting motion ($v_i$). To collect these data, the control input can be gradually varied from zero to its nominal positive and negative values, while the corresponding wheel velocities are recorded from the sensors. \textit{Appendix A} provides the core MATLAB code structure used to implement the proposed DNN policy.

\subsection{Architecture and Notation}
The feedforward DNN architecture comprises multiple layers of interconnected computational units (neurons), each applying a linear transformation followed by a nonlinear activation function \cite{saha2023adu}. In the present study, the linear velocity of each side of the SSWMR, denoted by $v_i$, serves as the sole input to the corresponding side's network. The output of the network for each side is a single scalar value representing the rpm velocity of the PMSM, denoted by $n_{p_i}$, which functions as the control command.
There are $L \in \mathbb{R}^+$ hidden layers, where the $\ell$-th hidden layer comprises $n_{\ell}$ neurons, with $\ell = 1, 2, \ldots, L$. The output layer corresponds to layer $L+1$--which in this work, consists of a single neuron. The weight matrix for layer $\ell$ is denoted by $W^{(\ell)} \in \mathbb{R}^{n_{\ell} \times n_{\ell-1}}$ and the corresponding bias vector by $b^{(\ell)} \in \mathbb{R}^{n_{\ell}}$.
For each layer $\ell$, the pre-activation (affine transformation) is represented by $z^{(\ell)} \in \mathbb{R}^{n_{\ell}}$ and the post-activation (i.e., the output of the layer) by $a^{(\ell)} \in \mathbb{R}^{n_{\ell}}$.
The input to the first layer is denoted as $a^{(0)} := v_i \in \mathbb{R}$. For each layer $\ell = 1, \ldots, L+1$, the pre-activation and activation are computed as
\begin{equation}
\begin{aligned}
\small
\label{10}
z^{(\ell)} = W^{(\ell)} a^{(\ell-1)} + b^{(\ell)}, \quad a^{(\ell)} = \varphi^{(\ell)}\left(z^{(\ell)}\right)
\end{aligned}
\end{equation}
where $\varphi^{(\ell)}$ denotes the (component-wise) activation function at layer $\ell$. For the hidden layers, we adopt the hyperbolic tangent (Tanh) activation function as $\varphi(z) = \tanh(z)$, where $\varphi(z) \in (-1, 1)$ \cite{shakiba2020novel}. The regression (i.e., linear activation) function used for the output layer is
$
\varphi^{(L+1)}\left(z^{(L+1)}\right)=z^{(L+1)}
$. Thus, the final output of the network is
$
{n}_{p_i}=a^{(L+1)}=z^{(L+1)} \in \mathbb{R}
$.
This allows the network to produce real-valued outputs, which are suitable for controlling the multi-stage actuation system.

\subsection{Forward Propagation}
Given the input scalar $v_{i}$, the network output is computed sequentially across layers as follows. The first hidden layer is obtained as

\begin{equation}
\begin{aligned}
\small
\label{11}
z^{(1)}=W^{(1)} v_{i}+b^{(1)}, \quad a^{(1)}=\varphi^{(1)}\left(z^{(1)}\right)
\end{aligned}
\end{equation}
Then, the subsequent hidden layers $(\ell=2, \ldots, L)$ are calculated as
\begin{equation}
\begin{aligned}
\small
\label{12}
z^{(\ell)}=W^{(\ell)} a^{(\ell-1)}+b^{(\ell)}, \quad a^{(\ell)}=\varphi^{(\ell)}\left(z^{(\ell)}\right)
\end{aligned}
\end{equation}
Finally, the output layer $(\ell=L+1)$ is achieved as
\begin{equation}
\begin{aligned}
\small
\label{13}
z^{(L+1)}=W^{(L+1)} a^{(L)}+b^{(L+1)}, \quad a^{(L+1)}=z^{(L+1) = {n}_{p_i}}
\end{aligned}
\end{equation}
When processing a batch of $P$ samples, the inputs are arranged as columns of a matrix:
\begin{equation}
\begin{aligned}
\small
\label{14}
\mathcal{V}_{i}=\left[v_{i}^{(1)}\left|v_{i}^{(2)}\right| \cdots \mid v_{i}^{(P)}\right] \in \mathbb{R}^{1 \times P}
\end{aligned}
\end{equation}
In matrix form, the layer outputs are computed as
\begin{equation}
\begin{aligned}
\small
\label{15}
A^{(\ell)}=\varphi^{(\ell)}\left(W^{(\ell)} A^{(\ell-1)}+b^{(\ell)} \mathbf{1}^T\right), \quad A^{(0)}=\mathcal{V}_{i}
\end{aligned}
\end{equation}

where $\mathbf{1} \in \mathbb{R}^P$ is a column vector of ones, and $\varphi^{(\ell)}$ is applied element-wise to its argument.

\subsection{Loss (Cost) Functions}
Since we use regression tasks, to assess the agreement between the network's predictions ${n}_{p_i}$ and the true targets $t_{p_i}$, we introduce a loss function $E_{\mathrm{MSE}}(\theta)$ based on mean squared error (MSE), where $\theta=\left\{W^{(\ell)}, b^{(\ell)}\right\}_{\ell=1}^{L+1}$ denotes the collection of all weights and biases. $E_{\mathrm{MSE}}(\theta)$ is defined as

\begin{equation}
\begin{aligned}
\small
\label{16}
E_{\mathrm{MSE}}(\theta)=\frac{1}{P} \sum_{p=1}^P\left\|{n}_{p_i}^{(p)}-t_{p_i}^{(p)}\right\|_2^2
\end{aligned}
\end{equation}

Equivalently, if we stack all network outputs and targets into matrices ${N}_{p_i}, T_{p_i} \in \mathbb{R}^{n_{L+1} \times P}$, then

\begin{equation}
\begin{aligned}
\small
\label{17}
E_{\mathrm{MSE}}(\theta)=\frac{1}{P}\|{N}_{p_i}-T_{p_i}\|_F^2
\end{aligned}
\end{equation}
where $\|\cdot\|_F$ denotes the Frobenius norm.

\subsection{Levenberg–Marquardt Backpropagation}
Although deep‐learning frameworks typically employ first‐order (stochastic) optimization algorithms, moderate‐sized networks may be trained using the second‐order‐inspired Levenberg–Marquardt (LM) backpropagation method \cite{zhou2018levenberg, almalki2020levenberg}. LM formulates training as a nonlinear least‐squares problem and approximates the Hessian via the Gauss–Newton approach, augmented by a damping parameter. As a result, it often converges in fewer iterations than conventional gradient descent.
For a network with parameter vector $w \in \mathbb{R}^{N_w}$ (stacking all weights and biases), let
\begin{equation}
\begin{aligned}
\small
\label{18}
\xi_{p_i}^{(p)}={n}_{p_i}^{(p)}(w)-t_{p_i}^{(p)} \quad \in \mathbb{R}^{n_{L+1}}
\end{aligned}
\end{equation}
be the error for sample $p$. Note that the vector $w \in \mathbb{R}^{N_w}$ collects each trainable parameter in the network-namely, all entries of each weigh matrix $W^{(\ell)}$ and each bias vector $b^{(\ell)}$ for $\ell=1, \ldots, L+1$. Concretely, if layer $\ell$ has
$W^{(\ell)} \in \mathbb{R}^{n_{\ell} \times n_{\ell-1}} \quad \text { and } \quad b^{(\ell)} \in \mathbb{R}^{n_{\ell}}$, then all $n_{\ell} \times n_{\ell-1}$ components of $W^{(\ell)}$ and all $n_{\ell}$ components of $b^{(\ell)}$ are "flattened" and concatenated into a single long vector. The total length

\begin{equation}
\begin{aligned}
\small
\label{19}
N_w=\sum_{\ell=1}^{L+1}\left(n_{\ell} n_{\ell-1}+n_{\ell}\right)
\end{aligned}
\end{equation}
is simply the sum over the layers of a fully connected neural network, where each term represents the number of weights in $W^{(\ell)}$ plus the number of biases in $b^{(\ell)}$. Training then optimizes this vector $w$ to minimize the chosen loss. Stacking all errors into a single vector yields
\begin{equation}
\begin{aligned}
\small
\label{20}
\xi_{p_i}(w)=\left[\begin{array}{c}
\xi_{p_i}^{(1)} \\
\xi_{p_i}^{(2)} \\
\vdots \\
\xi_{p_i}^{(P)}
\end{array}\right] \in \mathbb{R}^{m P}, \quad m:=n_{L+1}=1
\end{aligned}
\end{equation}
Then, the total MSE (scaled by $1 / P$ ) is
\begin{equation}
\begin{aligned}
\small
\label{21}
E_{\mathrm{MSE}}(w)=\frac{1}{P}\|\xi_{p_i}(w)\|_2^2=\frac{1}{P} \xi_{p_i}(w)^T \xi_{p_i}(w)
\end{aligned}
\end{equation}
Finding the weights $w$ that minimize $E_{\mathrm{MSE}}(w)$ is a nonlinear least-squares problem.

\subsection{Jacobian and Gauss-Newton Approximation}
The Jacobian $J(w) \in \mathbb{R}^{(m P) \times N_w}$ is defined as
\begin{equation}
\begin{aligned}
\small
\label{22}
J_{k, j}(w)=\frac{\partial s_k(w)}{\partial w_j}
\end{aligned}
\end{equation}

where $s_k$ is the $k$-th component of the stacked error vector $\xi_{p_i}$. In other words, each row of $J$ corresponds to $\partial s_k / \partial w$. The exact gradient of $E_{\mathrm{MSE}}$ is
\begin{equation}
\begin{aligned}
\small
\label{23}
\nabla E_{\mathrm{MSE}}(w)=\frac{2}{P} J(w)^T \xi_{p_i}(w)
\end{aligned}
\end{equation}
The exact Hessian is defined as
\begin{equation}
\begin{aligned}
\small
\label{24}
H(w)&=\nabla^2 E_{\mathrm{MSE}}(w)\\
&=\frac{2}{P}\left(J(w)^T J(w)+\sum_{k=1}^{m P} s_k(w) \nabla^2 s_k(w)\right)
\end{aligned}
\end{equation}
In Gauss-Newton, one neglects the second term (which involves second derivatives of the model outputs). Thus,
\begin{equation}
\begin{aligned}
\small
\label{25}
&H(w) \approx \frac{2}{P} J(w)^T J(w)\\
&\nabla E_{\mathrm{MSE}}(w)=\frac{2}{P} J(w)^T \xi_{p_i}(w)
\end{aligned}
\end{equation}

Solving the exact Newton step $\Delta w=-H(w)^{-1} \nabla E(w)$ is expensive (and requires computing second derivatives). Gauss-Newton approximates $H$ by $J^T J$, giving a step
\begin{equation}
\begin{aligned}
\small
\label{26}
\Delta w_{\mathrm{GN}}=-\left(J^T J\right)^{-1} J^T \xi_{p_i}
\end{aligned}
\end{equation}
However, $J^T J$ can still be near-singular or ill-conditioned, leading to unstable updates. Hence, LM can add a damping Tikhonov regularization term $\mu I$ to $J^T J$. The LM step $\Delta w$ solves
\begin{equation}
\begin{aligned}
\small
\label{27}
\left(J^T J+\mu I\right) \Delta w=-J^T \xi_{p_i}
\end{aligned}
\end{equation}
Equivalently,
\begin{equation}
\begin{aligned}
\small
\label{28}
\Delta w=-\left(J^T J+\mu I\right)^{-1} J^T \xi_{p_i}
\end{aligned}
\end{equation}
When $\mu$ is large, the term $\mu I$ dominates and
\begin{equation}
\begin{aligned}
\small
\label{29}
\Delta w \approx-\frac{1}{\mu} J^T \xi_{p_i}=-\eta \nabla E_\mathrm{MSE}, \quad \eta=\frac{1}{\mu}
\end{aligned}
\end{equation}
i.e., a small gradient-descent step. When $\mu$ is small ($\mu \rightarrow 0$), one recovers the Gauss-Newton step.
Thus, LM adaptively interpolates between gradient descent (stable but slow) and Gauss-Newton (fast but potentially unstable).
For damping‐parameter adjustment, at each iteration $b$, first, $J_b=J\left(w_b\right)$ and $\xi_b=\xi\left(w_b\right)$ are computed; second, $\mu_b$ is chosen (initially small); third,
\begin{equation}
\begin{aligned}
\small
\label{30}
\left(J_b^T J_b+\mu_b I\right) \Delta w_b=-J_b^T \xi_b
\end{aligned}
\end{equation}
 is solved; fourth, the candidate new weights $w_{\text {cand }}=w_b+\Delta w_b$ are evaluated; finally, the new error $E_\mathrm{MSE}\left(w_{\text {cand }}\right)$ is computed. If $E_\mathrm{MSE}\left(w_{\text {cand }}\right)<E_\mathrm{MSE}\left(w_b\right)$, the step $w_{b+1} \leftarrow w_{\text {cand }}$ is accepted and $\mu_{b+1}=\mu_b / \beta$ (commonly $\beta=10$ or a similar factor) is decreased. If $E_\mathrm{MSE}\left(w_{\text {cand }}\right) \geq E\left(w_b\right)$ the step $w_{b+1}=w_b$ is rejected, $\mu_{b+1}=\mu_b \times \beta$ is increased.
By reducing $\mu$ when the error is decreasing, LM approaches pure Gauss-Newton, accelerating convergence near a local minimum. By increasing $\mu$ when the step fails, it behaves more like gradient descent, ensuring stability. To illustrate the LM training for the DNN-based control policy of the SSWMR, the structure of \textbf{Algorithm 1} is presented.
\begin{table}[h]
  \centering
  \begin{threeparttable}
  \renewcommand{\arraystretch}{1.0}
  \begin{tabular}{p{0.95\linewidth}}
    \hline
    \\
    \multicolumn{1}{c}{\textbf{Algorithm 1}: LM training for DNN-based control policy} \\
    \\
    \hspace{0.4cm}\textbf{Input}: Initial weights $w_0$, damping factor $\mu_0$, multiplier $\beta > 1$, tolerance $\epsilon$, training data $\{(v_i^{(p)}, t_{p_i}^{(p)})\}_{p=1}^P$ \\
    \hspace{0.4cm}\textbf{Output}: Trained weights $w$ \\
    \\
    {\small 1}\hspace{0.3cm}Set $w \leftarrow w_0$, $\mu \leftarrow \mu_0$ \\
    {\small 2}\hspace{0.3cm}\textbf{repeat} \\
    {\small 3}\hspace{0.7cm}Compute network outputs $\hat{n}_{p_i}^{(p)}$ and errors $\xi^{(p)} = \hat{n}_{p_i}^{(p)} - t_{p_i}^{(p)}$ \\
    {\small 4}\hspace{0.7cm}Stack all errors into vector $\xi(w) \in \mathbb{R}^{P}$ \\
    {\small 5}\hspace{0.7cm}Compute Jacobian matrix $J(w) \in \mathbb{R}^{P \times N_w}$ \\
    {\small 6}\hspace{0.7cm}Solve: $(J^\top J + \mu I)\Delta w = -J^\top \xi(w)$ \\
    {\small 7}\hspace{0.7cm}Compute candidate: $w_{\text{cand}} = w + \Delta w$ \\
    {\small 8}\hspace{0.7cm}Evaluate $E_{\text{cand}} = E_{\mathrm{MSE}}(w_{\text{cand}})$ \\
    {\small 9}\hspace{0.7cm}\textbf{if} $E_{\text{cand}} < E_{\mathrm{MSE}}(w)$ \textbf{then} \\
    {\small 10}\hspace{1.0cm}Accept step: $w \leftarrow w_{\text{cand}}$ \\
    {\small 11}\hspace{1.0cm}Decrease damping: $\mu \leftarrow \mu / \beta$ \\
    {\small 12}\hspace{0.5cm}\textbf{else} \\
    {\small 13}\hspace{1.0cm}Reject step: $w$ unchanged \\
    {\small 14}\hspace{1.0cm}Increase damping: $\mu \leftarrow \mu \cdot \beta$ \\
    {\small 15}\hspace{0.2cm}\textbf{until} $\|\Delta w\| < \epsilon$ or max iterations reached \\
    {\small 16}\hspace{0.2cm}\textbf{return} $w$ \\
    \\
    \hline
  \end{tabular}
    \begin{tablenotes}
      \item[- The Jacobian $J(w)$ contains partial derivatives of output error with respect to weights.] 
      \item[- All operations are performed on mini-batches or full datasets depending on the training setup.]
    \end{tablenotes}
  \end{threeparttable}
\end{table}

\begin{figure*}[h!]
\hspace*{-0.0cm} 
\centering
\includegraphics[width=0.75\textwidth, height=5cm]{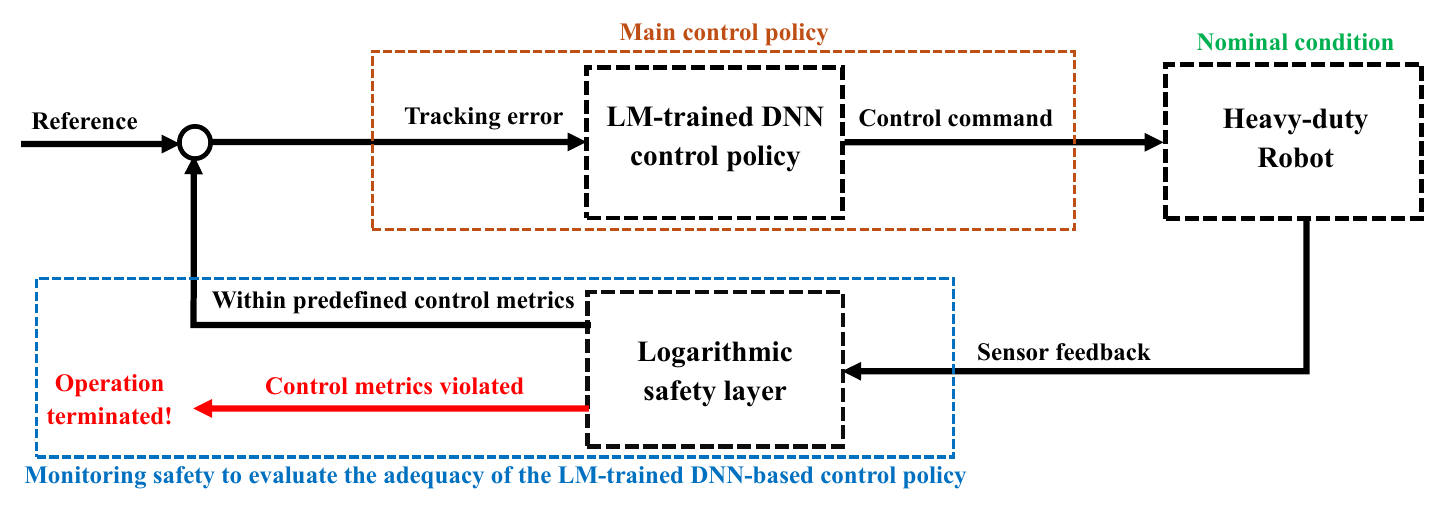}
\caption{The schematic of the proposed safe LM-trained DNN-based control policy for the SSWMR in nominal condition.}
\label{i9so_LM_dnn}
\end{figure*}

\subsection{Safety Layer Design for DNN-based Control Policy}
Upon completion of training, the resulting model parameters encapsulate a data-driven control policy that maps reference wheel velocities $v_{{ref}_i}$ of the SSWMR to PMSM control commands $n_{p_i}$ with high fidelity under nominal conditions. Note that the DNN is trained using actual system output data $v_{i}$, while in operation, the ideal velocity reference $v_{{ref}_i}$ is passed to the encapsulated DNN-based controller to produce the appropriate control commands. To ensure high-level safety in the SSWMR system with the proposed LLM-trained DNN-based control policy, we employ the PPC framework to monitor and regulate critical metrics of the system response, such as overshoot and steady-state error. In PPC, key elements such as singularities and zeros are strategically employed to impose constraints that keep the system's state within a desired performance range. This is typically achieved by embedding a barrier function or transformation within the control law, effectively preventing the state from reaching or exceeding undesirable limits \cite{wang2023concurrent, gfdwsfdfd21prague}. Following the approach in \cite{zhang2019low, shahna2025model}, the tracking error metrics for each side of the robot using DNN-based controllers can be constrained within an exponential, symmetric, time-varying interval defined by $-\zeta(t)<e_i<\zeta(t)$, where:

\begin{equation}
\begin{aligned}
\small
\label{31}
\zeta(t)=\left(\zeta^{shoot}-\zeta^{bound}\right) e^{-\zeta^* t}+\zeta^{bound}
\end{aligned}
\end{equation}

Given that $\zeta^{{shoot }}>\zeta^{{bound }}>0$ and $\zeta_i^*>0$, the overshoot of $e_i$ is constrained to remain below $\zeta^{{shoot }}$. The parameters $\zeta^{{bound }}$ and $\zeta_i^*$ define the steady-state bound and the convergence rate of $e_i$, respectively. Let us define a conditional safety-monitoring function for each side of the SSWMR based on the expression $\log \left(\frac{\zeta^2}{\zeta^2-e_i^2}\right)$, whose domain is restricted by the requirement that the denominator remains positive. If the initial error satisfies $e_i\left(t_0\right)<\zeta\left(t_0\right)$, then as $e_i$ approaches $\zeta$, the denominator approaches zero, causing the logarithmic term to diverge toward infinity. This results in a singularity, rendering the function undefined due to a non-positive argument within the logarithm. This conditional safety-monitoring mechanism can be readily implemented in real-time environments, where execution can be halted upon detection of warnings such as ``Infinity or NaN value encountered" when the computed value exceeds numerical limits. This behavior is beneficial for ensuring that the tracking error does not exceed the prescribed bound $\zeta$, thereby preventing potentially unsafe or undesirable system responses in a control setting. The related MATLAB code for this conditional safety-monitoring mechanism is provided in \textit{Appendix B}.

Figure \ref{i9so_LM_dnn} illustrates the standard schematic of the proposed LM-trained DNN-based control policy generating control commands for a heavy-duty robot, based on tracking error. To ensure safety, the logarithmic monitoring layer evaluates whether the control performance remains within PPC bounds and halts operation if violations are detected. Although the safety layer ensures that the system remains within a predefined safe operating region, it acts passively—terminating operation upon boundary violation—without actively guiding the system back toward safer states as the limits are approached, leading to conservatism in the resulting safe sets.  To illustrate the LM-trained DNN control policy for the SSWMR, the structure of \textbf{Algorithm 2} is presented.

\begin{table}[h]
  \centering
  \begin{threeparttable}
  \renewcommand{\arraystretch}{1.0}
  \begin{tabular}{p{0.95\linewidth}}
    \hline
    \\
    \multicolumn{1}{c}{\textbf{Algorithm 2}: Safe LM-trained DNN policy for each side of the SSWMR} \\
    \\
    \hspace{0.4cm}\textbf{Input}: PPC metrics for both safety layers $\zeta^{\text{shoot}}$, $\zeta^{\text{bound}}$, $\zeta^*$, reference velocities $v_{\text{ref}_i}$, sensor feedback $v_i$\\
    \hspace{0.4cm}\textbf{Output}: Control command $n_{p_i}$ \\
    \\
    {\small 1}\hspace{0.3cm}\textbf{for each side } $i$ \textbf{do} \\
    {\small 2}\hspace{0.7cm}Compute tracking error: $e_i = v_i - v_{\text{ref}_i}$ \\
    {\small 3}\hspace{0.7cm}Compute low-level safety: $\zeta = \left(\zeta^{\text{shoot}} - \zeta^{\text{bound}}\right)e^{-\zeta^* t} + \zeta^{\text{bound}}$ \\
    {\small 4}\hspace{0.7cm}Compute denominator: $\text{denom} = \zeta^2 - e_i^2$ \\
    {\small 5}\hspace{0.7cm}\textbf{if} $\text{denom} \leq 0$ \textbf{then} \\
    {\small 6}\hspace{1.0cm}Raise safety violation and go to \textbf{step 11}  \\
    {\small 7}\hspace{0.7cm}\textbf{else} \\
    {\small 8}\hspace{1cm}Pass $v_{\text{ref}_i}$ to LM-trained DNN policy to compute $n_{p_i}$ \\
    {\small 9}\hspace{1cm}Return to \textbf{step 3} \\
    {\small 10}\hspace{0.6cm}\textbf{end if} \\
    {\small 11}\hspace{0.6cm}\textbf{terminate operation}\\
    {\small 12}\hspace{0.2cm}\textbf{end for} \\
    \\
    \hline
  \end{tabular}
    \begin{tablenotes}
      \item[- The PPC metrics are manually adjustable for use in a wide range of application scenarios.]
      \item[- If safety condition is violated, $n_{p_i}$ is not generated to protect the system]
    \end{tablenotes}
  \end{threeparttable}
\end{table}

\section{Safe Model-Free RAC Policy}
\label{section:state_space_EMLA}

In this section, we develop a stability-guaranteed RAC strategy for the SSWMR system under study. In control theory, a fundamental trade-off exists between robustness and responsiveness—improving one often degrades the other. This balance is particularly critical in the presence of external disturbances, where enhanced robustness can ensure continued safe operation. Acknowledging this trade-off, we propose a novel adaptive control policy for each side of the system, specifically designed to improve robustness without compromising overall stability. In PPC, key elements such as singularities and zeros are strategically employed to impose constraints that ensure the system's state remains within a desired performance range. Following the approach in \cite{tee2009barrier}, let us define an adaptive law for each side, as
\begin{equation}
\begin{aligned}
\small
\label{32}
\dot{\hat{\theta}}_i = -\delta_i \hat{\theta}_i+\gamma_i   (\frac{{e}_i}{o^2-{e}_i^2})^2
\end{aligned}    
\end{equation}
where $\hat{\theta}_i(t_0)$, $\delta_i$, and $\gamma_i$ are positive constants. Hence, the tracking error for each side of the robot is constrained within a symmetric, time-varying interval defined by $-o(t)<e_i<o(t)$. Simialr to Eq. \eqref{31}, we can define a decaying exponential bounds for control error performance, as
\begin{equation}
\begin{aligned}
\small
\label{33}
o(t)=\left(o^{shoot}-o^{bound}\right) e^{-o^* t}+o^{bound}
\end{aligned}
\end{equation}
where $o^{\text {shoot }}>o^{\text {bound }}>0$ and $o^*>0$. This formulation ensures that the initial error does not exceed $o^{\text {shoot }}$, while the long-term tracking performance is bounded by $o^{\text {bound }}$ with a convergence rate governed by $o^*$. Assume that the adaptive parameter $\hat{\theta}_i$ estimates parameter $\theta^*_i$. Thus, we can define the adaptive error $\tilde{\theta}_i = \hat{\theta}_i -\theta^*_i$. From Eq. \eqref{32}, we have
\begin{equation}
\begin{aligned}
\small
\label{34}
\dot{\tilde{\theta}}_i = -\delta_i \tilde{\theta}_i+\gamma_i   (\frac{{e}_i}{o_i^2-{e}_i^2})^2 - \delta_i \theta^*_i
\end{aligned}    
\end{equation}
We can define a logarithmic quadratic function, as 
\begin{equation}
\begin{aligned}
\small
\label{35}
V_i=\frac{1}{2} \log \left(\frac{o^2}{o^2 - e^2_i}\right)+ \frac{1}{2} {A_i}\tilde{\theta}_i^2
\end{aligned}
\end{equation}
This logarithmic barrier function serves as the safety metric and can be implemented in MATLAB as shown in \textit{Appendix D}. After differentiating \eqref{35}, we have
\begin{equation}
\begin{aligned}
\small
\label{36}
\dot{V}_i=\frac{-\dot{o} e_i^2+e_i o \dot{e}_i}{o\left(o^2-e_i^2\right)}+A_i \tilde{\theta}_i \dot{\tilde{\theta}}_i
\end{aligned}
\end{equation}
By inserting \eqref{9} into \eqref{36}, we have
\begin{equation}
\begin{aligned}
\small
\label{37}
\dot{V}_i= & \frac{-\dot{o} e_i^2}{o\left(o^2-e_i^2\right)} \\
&+\frac{e_i}{o^2-e_i^2}[A_i n_{p_i} +\tau^{-1} (F_i - v_i)- \dot{v}_{{ref}_i}]\\
&+A_i \tilde{\theta}_i \dot{\tilde{\theta}}_i
\end{aligned}
\end{equation}
we can obtain
\begin{equation}
\begin{aligned}
\small
\label{38}
\dot{V}_i= & \frac{-\dot{o}_i e_i^2}{o\left(o^2-e_i^2\right)}+A_i\frac{e_i}{o^2-e_i^2} n_{p_i} \\
&+\frac{e_i}{o^2-e_i^2}\tau^{-1} (F_i - v_i)  - \frac{e_i}{o_i^2-e_i^2}\dot{v}_{{ref}_i}+A_i \tilde{\theta}_i \dot{\tilde{\theta}}_i
\end{aligned}
\end{equation}
Now, we propose the RAC policy for the PMSM control command $n_{p_i}$ as 
\begin{equation}
\begin{aligned}
\small
\label{39}
n_{p_i} = -  \frac{1}{2} k_i e_i - \gamma_i  \frac{e_i}{o^2-e_i^2} \hat{\theta}_i
\end{aligned}
\end{equation}
Inserting \eqref{39} into \eqref{38}, we have
\begin{equation}
\begin{aligned}
\small
\label{40}
\dot{V}_i \leq & A_i\frac{e_i}{o^2-e_i^2} (-  \frac{1}{2} k_i e_i - \gamma_i  \frac{e_i}{o^2-e_i^2} \hat{\theta}_i)\\
&+|\frac{e_i}{o^2-e_i^2}|\hspace{0.1cm}|\hspace{0.1cm}\tau^{-1} (F_i - v_i) - \dot{v}_{{ref}_i}-\frac{\dot{o}\hspace{0.1cm} {e_i}}{o}\hspace{0.1cm}|\\
& +A_i \tilde{\theta}_i \dot{\tilde{\theta}}_i
\end{aligned}
\end{equation}
Considering the proposed adaptive law error defined in \eqref{34}, we obtain
\begin{equation}
\begin{aligned}
\small
\label{41}
\dot{V}_i \leq & - \frac{1}{2} {A_i} k_i \frac{{e}^2_i}{o^2-e_i^2}  - {A_i} \gamma_i  (\frac{{e}_i}{o^2-e_i^2})^2 \hat{\theta}_i  \\
&+  |\frac{{e}_i}{o^2-e_i^2}| \hspace{0.2cm} \hspace{0.1cm}|\hspace{0.1cm}\tau^{-1} (F_i - v_i) - \dot{v}_{{ref}_i}-\frac{\dot{o}\hspace{0.1cm} {e_i}}{o}\hspace{0.1cm}| \\
&-\delta_i {A_i} \tilde{\theta}^2_i+\gamma_i  {A_i} (\frac{\bar{e}_i}{o^2-e_i^2})^2  \tilde{\theta}_i - {A_i}\delta_i \theta^*_i \tilde{\theta}_i 
\end{aligned}
\end{equation}
Based on \textit{Assumption II.3}, we can assume that there exists a positive parameter $f^*_i$ such that
\begin{equation}
\begin{aligned}
\small
\label{42}
|\hspace{0.1cm}\tau^{-1} (F_i - v_i) - \dot{v}_{{ref}_i}-\frac{\dot{o}\hspace{0.1cm} {e_i}}{o}\hspace{0.1cm}| \leq f^*_i
\end{aligned}
\end{equation}

\textit{Remark IV.1.} By applying the Cauchy-Schwarz inequality, one can establish that for any scalars $a$ and $b$, the product satisfies $a b \leq \epsilon_i a^2+\frac{1}{4 \epsilon_i} b^2$, where $\epsilon_i>0$ is an arbitrary positive constant.

\begin{figure*}[h!]
\hspace*{-0.0cm} 
\centering
\includegraphics[width=0.75\textwidth, height=6cm]{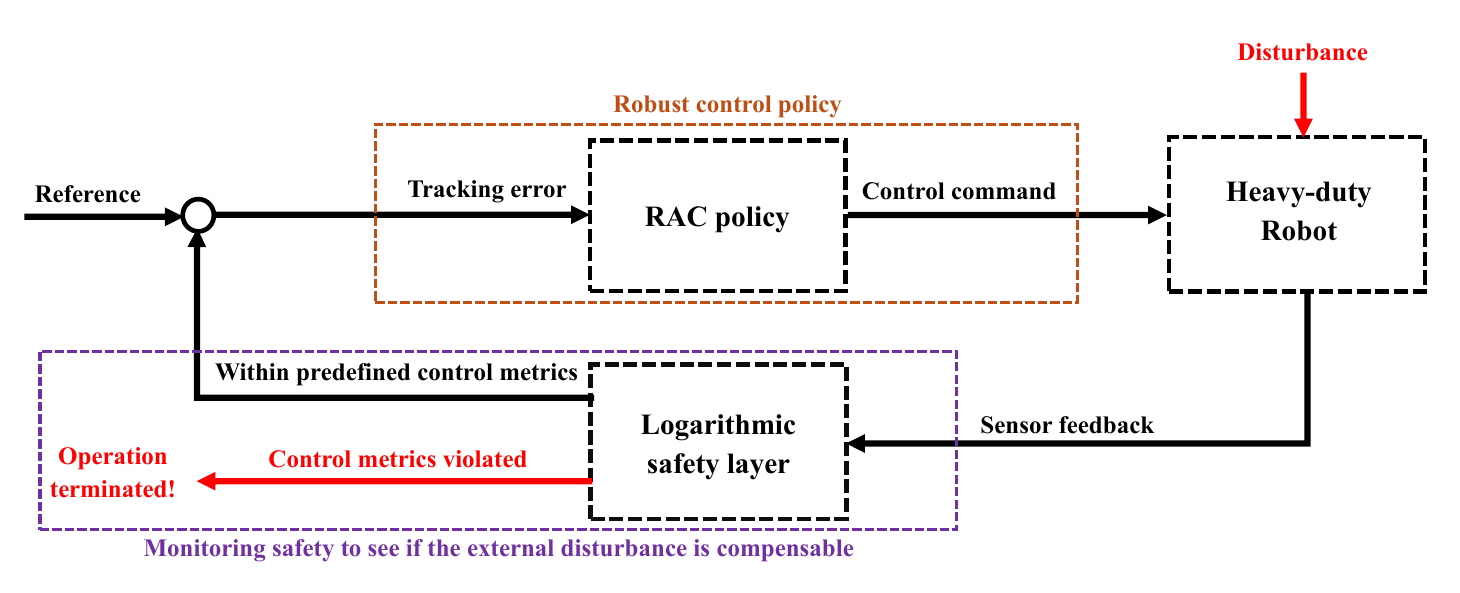}
\caption{The schematic of the proposed safe stability-guaranteed RAC policy for the SSWMR under disturbance.}
\label{i9so_stale_arac}
\end{figure*}

Thus, from \textit{Remark IV.1}, \eqref{41} and \eqref{42}, as well as knowing $\tilde{\theta}_i = \hat{\theta}_i -\theta^*_i$, we have
\begin{equation}
\begin{aligned}
\small
\label{43}
\dot{V}_i \leq & - \frac{1}{2} {A_i} k_i \frac{{e}^2_i}{o^2-e_i^2}  - {A_i} \gamma_i  (\frac{{e}_i}{o^2-e_i^2})^2 {\theta}^*_i   \\
&+  \epsilon_i (\frac{{e}_i}{o^2-e_i^2})^2 {f^*_i}^2 + \frac{1}{4 \epsilon_i}-\delta_i {A_i} \tilde{\theta}^2_i- {A_i}\delta_i \theta^*_i \tilde{\theta}_i 
\end{aligned}
\end{equation}
Let us define unknown positive $\theta^*_i$ as
\begin{equation}
\begin{aligned}
\small
\label{44}
\theta^*_i = \frac{\epsilon_i {f^*_i}^2}{\gamma_i A_i}
\end{aligned}    
\end{equation}
Thus, we can reach
\begin{equation}
\begin{aligned}
\small
\label{45}
\dot{V}_i \leq & - \frac{1}{2} {A_i} k_i \frac{{e}^2_i}{o^2-e_i^2} + \frac{1}{4 \epsilon_i} \\
&-\delta_i {A_i} \tilde{\theta}^2_i- {A_i}\delta_i \theta^*_i \tilde{\theta}_i 
\end{aligned}
\end{equation}
{\textit{Lemma IV.1.} For any positive parameter $o$, the following inequality holds for all ${e}_i$ satisfying $\left|{e}_i\right|<o$ \cite{ren2010adaptive}:
\begin{equation}
\begin{aligned}
\small
\label{46}
\log \left(\frac{o^2}{o^2-e_i^2}\right)<\frac{{e}_i^2}{o^2-e_i^2}
\end{aligned}
\end{equation}}
Thus, from \textit{Lemma IV.1} and \eqref{45}, we have
\begin{equation}
\begin{aligned}
\small
\label{47}
\dot{V}_i \leq & - \frac{1}{2} {A_i} k_i \log \left(\frac{o_i^2}{o^2-e_i^2}\right) + \frac{1}{4 \epsilon_i}-\delta_i {A_i} \tilde{\theta}^2_i \\
&- {A_i}\delta_i \theta^*_i \tilde{\theta}_i 
\end{aligned}
\end{equation}
Evidently,
\begin{equation}
\begin{aligned}
\small
\label{48}
\dot{V}_i \leq & - \frac{1}{2} {A_i} k_i  \log \left(\frac{o_i^2}{o^2-e_i^2}\right)  + \frac{1}{4 \epsilon_i} -\frac{1}{2}\delta_i {A_i} \tilde{\theta}^2_i \\
& - \frac{1}{2} {A_i} {\delta}_i (\hat{\theta}_i - \theta^*_i)^2 - {A_i} {\delta}_i \theta^*_i (\hat{\theta}_i - \theta^*_i)
\end{aligned}
\end{equation}
After simplifications, we have
\begin{equation}
\begin{aligned}
\small
\label{49}
\dot{V}_i \leq & - \frac{1}{2} {A_i} k_i  \log \left(\frac{o_i^2}{o^2-e_i^2}\right)  + \frac{1}{4 \epsilon_i} -\frac{1}{2}\delta_i {A_i} \tilde{\theta}^2_i \\
& + \frac{1}{2} A_i {\delta_i \theta^*_i}^2 
\end{aligned}
\end{equation}
we have
\begin{equation}
\begin{aligned}
\small
\label{50}
\dot{V}_i \leq & - \mu_i V_i + \ell_i 
\end{aligned}    
\end{equation}
where $\mu_i=\min[A_i k_i, \delta_i]$ and $\ell_i = \frac{1}{4 \epsilon_i} + \frac{1}{2} A_i {\delta_i \theta^*_i}^2 $. Thus, based on the resulting Eq. \eqref{50} and \cite{shahna2025robudfdsfst, shahasaasca2024asdacvrobustness, shahna2024exponential, corless1993bounded, shahna2024integrating, wu2020robust}, the proposed safe RAC policy for the SSWMR system provided in Eq. \eqref{9} is uniformly exponentially stable even in the presence of external disturbances. 

\textit{Remark IV.2.} The parameters $\ell_i$ plays a critical role in shaping the system's robustness and responsiveness. Specifically, $\ell_i$ is directly related to $\theta_i^*$, which characterizes the intensity of the external disturbance $f_i^*$ in Eqs. \eqref{42} and \eqref{44}. As evident from Eq. \eqref{50}, system stability is not guaranteed if $\ell_i$ becomes excessively large. This would violate the condition defined in \eqref{33}, resulting in a positive derivative of the Lyapunov function, thereby indicating instability. As the disturbance $f_i^*$ intensifies, the Lyapunov function value approaches zero. If the disturbance exceeds the compensatory capacity of the RAC, which is predefined based on actuator limitations, the system's robustness threshold is surpassed, prompting the safety layer to trigger a shutdown. To illustrate the safe RAC policy of the SSWMR, the structure of \textbf{Algorithm 3} is presented

\begin{table}[h]
  \centering
  \begin{threeparttable}
  \renewcommand{\arraystretch}{1.0}
  \begin{tabular}{p{0.95\linewidth}}
    \hline
    \\
    \multicolumn{1}{c}{\textbf{Algorithm 3}: Safe RAC policy for each side of the SSWMR} \\
    \\
    \hspace{0.4cm}\textbf{Input}: PPC metrics $o^{\text{shoot}}$, $o^{\text{bound}}$, $o^*$, reference velocities $v_{\text{ref}_i}$, sensor feedback $v_i$, design parameters $\delta_i$, $\gamma_i$, $k_i$ \\
    \hspace{0.4cm}\textbf{Output}: Control input commands $n_{p_i}$ \\
    \\
    {\small 1}\hspace{0.3cm}\textbf{for each side } $i$ \textbf{do} \\
    {\small 2}\hspace{0.7cm}Compute tracking error: $e_i = v_i - v_{\text{ref}_i}$ \\
    {\small 3}\hspace{0.7cm}Compute safety bound: $o = \left(o^{\text{shoot}} - o^{\text{bound}}\right)e^{-o^* t} + o^{\text{bound}}$ \\
    {\small 4}\hspace{0.7cm}Compute denominator: $\text{denom} = o^2 - e_i^2$ \\
    {\small 5}\hspace{0.7cm}\textbf{if} $\text{denom} \leq 0$ \textbf{then} \\
    {\small 6}\hspace{1.0cm}Raise safety violation and go to \textbf{step 12} \\
    {\small 7}\hspace{0.7cm}\textbf{else} \\
    {\small 8}\hspace{1.0cm}Compute safety metric: $\log\left(\frac{o^2}{o^2 - e_i^2}\right)$ \\
    {\small 9}\hspace{1.0cm}Update adaptive parameter: $\dot{\hat{\theta}}_i = -\delta_i \hat{\theta}_i + \gamma_i \left(\frac{e_i}{o^2 - e_i^2}\right)^2$ \\
    {\small 10}\hspace{0.85cm}Compute control input: $n_{p_i} = -\frac{1}{2}k_i e_i - \gamma_i \frac{e_i}{o(t)^2 - e_i^2} \hat{\theta}_i$ \\
    {\small 11}\hspace{0.55cm}\textbf{end if} \\
    {\small 12}\hspace{0.55cm}\textbf{terminate operation} \\
    {\small 13}\hspace{0.2cm}\textbf{end for} \\
    \\
    \hline
  \end{tabular}
    \begin{tablenotes}
         \item[- The PPC metrics are manually adjustable for use in a wide range of application scenarios.]
      \item[- If safety condition is violated, $n_{p_i}$ is not generated to protect the system.]
      \item[- Adaptive parameter $\hat{\theta}_i$ is updated using a logarithmic Lyapunov-based law.]
    \end{tablenotes}
  \end{threeparttable}
\end{table}

The related MATLAB code for this conditional safety-monitoring mechanism is provided in \textit{Appendix C}. Figure \ref{i9so_stale_arac} illustrates the standard schematic of the proposed stability-guaranteed RAC policy generating control commands for a heavy-duty robot, based on tracking error. To ensure safety, the logarithmic monitoring layer continuously monitors system performance during operation. It initiates a shutdown only when disturbances become sufficiently severe such that compensation is no longer viable and continued operation would jeopardize the system or its environment. 

\section{Synthesis Of Dnn And Rac Policy With Two Safety Layers}
\label{aasafadfgadA}
As previously noted, in practice, heavy-duty WMRs with complex actuation mechanisms operate predominantly under nominal conditions, where the proposed LM-trained DNN control policy can typically achieve high performance. These controllers may tolerate minor or transient external disturbances without significant performance degradation. However, to ensure safe and reliable operation—especially in rare but critical scenarios involving severe disturbances or faults—it is essential to guarantee robustness and system stability in compliance with stringent international safety standards.
In this section, we propose the integration of both control policies introduced in Sections \ref{section:trajectory_optimization} and \ref{section:state_space_EMLA}, augmented by two safety layers with differing levels of authority, to ensure continuous operation even in the presence of compensable external disturbances.
The framework operates primarily through the LM-trained DNN control policy, which provides high tracking accuracy under nominal conditions but offers limited robustness. During normal operation, this policy ensures excellent performance. However, its output is continuously monitored in real time by a low-level safety layer. If an external disturbance degrades performance beyond a predefined threshold, the system does not immediately terminate operation. Instead, a supervisory RAC module is activated, enabling the system to compensate for the disturbance and continue operating safely—though with reduced tracking accuracy for the remainder of the session. In this way, the low-level safety layer manages the inherent trade-off between robustness and responsiveness.
While the RAC policy delivers lower precision than the DNN controller, it is significantly more robust, allowing the system to operate reliably until scheduled maintenance or corrective action is taken. Notably, even while under RAC control, the system’s behavior continues to be monitored by the same safety layer. If the disturbance exceeds the RAC's compensatory capability—defined according to actuator limitations—the system’s robustness threshold is considered violated, prompting the safety layer to initiate a controlled shutdown. This prevents potential damage to the system, the environment, or surrounding assets.
To implement this fail-safe mechanism, we employ a latched function—a stateful construct that irreversibly manages the inherent trade-off between
system robustness and responsiveness by changing its output when a triggering condition is met. It is defined as follows:
\begin{equation}
\begin{aligned}
\small
\label{51}
u_{c_i} = \alpha_1 (t) {u_{\text{DNN}}}_i + \alpha_2 (t) u_{s_i}
\end{aligned}    
\end{equation} 
where $u_{s_i}$ and $u_{{DNN}_i}$ are the RAC and DNN control policy commands for $n_{p_i}$. Employing latched-OFF logic, we use $\alpha_1 (t)$ for the proposed DNN-based control that starts at 1 and switches to 0 when the condition (any external disturbance occurs that degrades performance beyond a predefined threshold) is met, and then remains off permanently. $\alpha_1 (t)$ is defined as
\begin{equation}
\begin{aligned}
\small
\label{52}
\alpha_1(t)= \begin{cases}1, & \text { if } e^2_i(\tau)<\zeta^2(\tau) \text { for all } \tau \in[0, t] \\ 0, & \text { if there exists } \tau \in[0, t] \text { such } \zeta^2(\tau) \leq e^2_i(\tau)\end{cases}
\end{aligned}    
\end{equation}
If we define the low-level safety layer margin as $R = \zeta^2 -e^2_i$, we can code $\alpha_1$ in MATLAB, as shown in \textit{Appendix E}. In contrast, we also use $\alpha_2$ for the proposed RAC policy in the condition the DNN control is not robust enough to maintain tracking performance. $\alpha_2$ first starts latched-ON logic function (starts at 0) and switches to 1 the moment the condition ($R \geq 0$) occurs, staying at 1 forever regardless of future inputs. $\alpha_2$ is defined as
\begin{equation}
\begin{aligned}
\small
\label{53}
\alpha_2(t)= \begin{cases}0, & \text { if } e^2_i(\tau)<\zeta^2(\tau) \text { for all } \tau \in[0, t] \\ 1, & \text { if there exists } \tau \in[0, t] \text { such } \zeta^2(\tau) \leq e^2(\tau)\end{cases}
\end{aligned}    
\end{equation}
We can code $\alpha_2$ in MATLAB, as shown in \textit{Appendix E}. To illustrate the synthesis of the LM-trained DNN control and the RAC policy for the SSWMR, the structure of \textbf{Algorithm 4} is presented
By using the proposed synthesis control policies for each side of the WMR, we can define the same quadratic function provided in \eqref{35} for analyzing the system stability from the switching time $t_s$ onward, as

\begin{figure*}[h!]
\hspace*{-0.0cm} 
\centering
\includegraphics[width=0.8\textwidth, height=10cm]{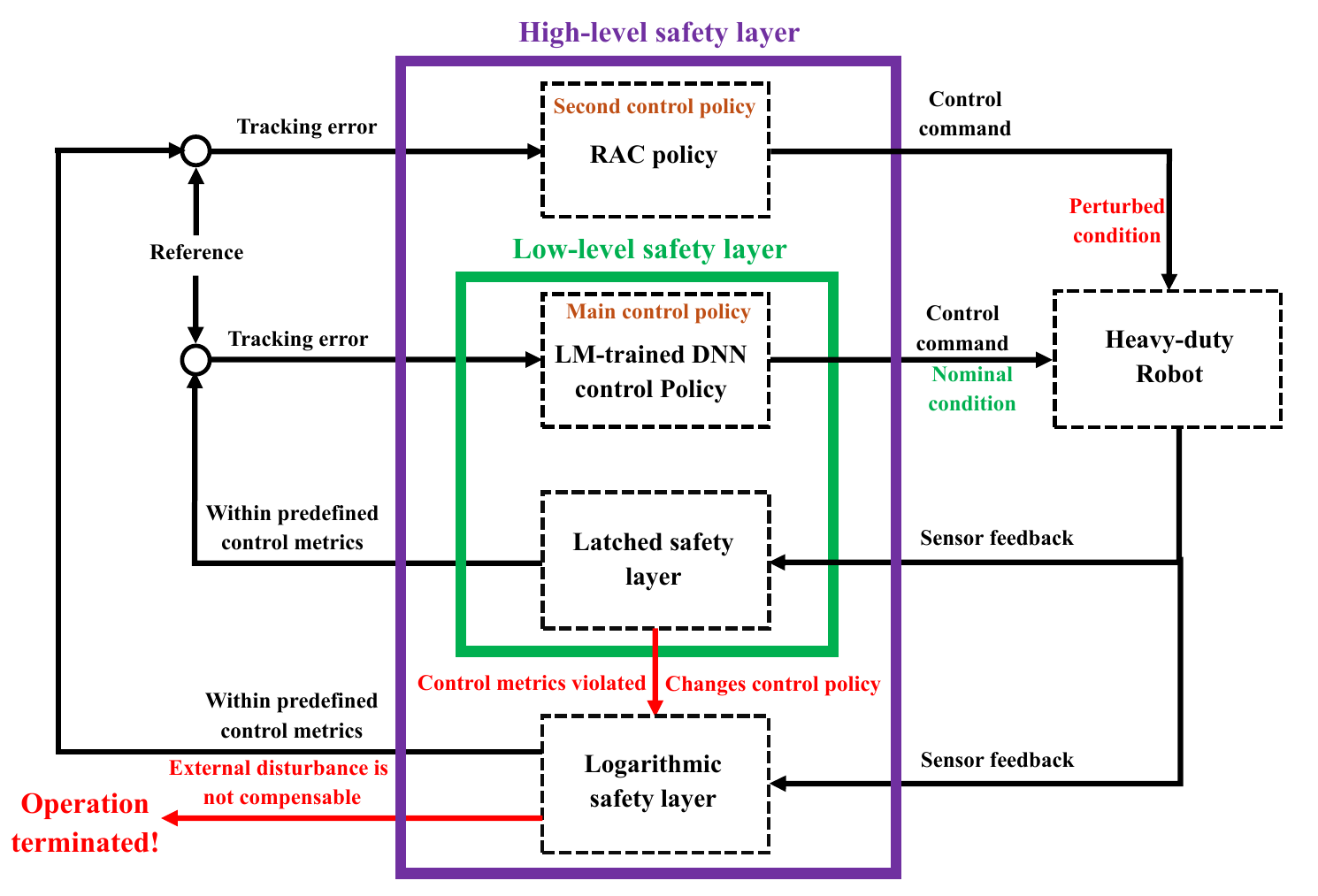}
\caption{The schematic of the synthesis of DNN and RAC policy for the SSWMR under both nominal and non-nominal conditions, monitored by two safety layers with different authorities.}
\label{i9sasddadado_synthes}
\end{figure*}

\begin{table}[h]
  \centering
  \begin{threeparttable}
  \renewcommand{\arraystretch}{1.0}
  \begin{tabular}{p{0.95\linewidth}}
    \hline
    \\
    \multicolumn{1}{c}{\textbf{Algorithm 4}: Safe synthesis control policy for each side of the SSWMR} \\
    \\
    \hspace{0.4cm}\textbf{Input}: PPC metrics for both safety layers $o$ and $\zeta$, reference velocities $v_{\text{ref}_i}$, sensor feedback $v_i$, design parameters $\delta_i$, $\gamma_i$, $k_i$\\
    \hspace{0.4cm}\textbf{Output}: Control command $u_{c_i} = n_{p_i} = \alpha_1 u_{{DNN}_i} + \alpha_2 u_{s_i}$ \\
    \\
    {\small 1}\hspace{0.3cm}\textbf{for each side } $i$ \textbf{do} \\
    {\small 2}\hspace{0.7cm}Initialize gating flags: $\alpha_1 \leftarrow 1$, $\alpha_2 \leftarrow 0$ (latch-on logic)\\
    {\small 3}\hspace{0.7cm}Compute tracking error: $e_i = v_i - v_{\text{ref}_i}$ \\
    {\small 4}\hspace{0.7cm}Compute low-level safety: $\zeta = \left(\zeta^{\text{shoot}} - \zeta^{\text{bound}}\right)e^{-\zeta^* t} + \zeta^{\text{bound}}$ \\
    {\small 5}\hspace{0.7cm}Compute $R = \zeta^2 - e_i^2$ \\
    {\small 6}\hspace{0.7cm}\textbf{if} $R \leq 0$ \textbf{then} (latch-off logic) \\
    {\small 7}\hspace{1.0cm}$\alpha_1 \leftarrow 0$, $\alpha_2 \leftarrow 1$ (switch to RAC policy) \\
    {\small 8}\hspace{1.0cm}Compute tracking error: $e_i = v_i - v_{\text{ref}_i}$ \\
    {\small 9}\hspace{1.0cm}Compute high-level safety:$o = \left(o^{\text{shoot}} - o^{\text{bound}}\right)e^{-o^* t} + o^{\text{bound}}$ \\
    {\small 10}\hspace{0.85cm}Compute denominator: $\text{denom} = o^2 - e_i^2$ \\
    {\small 11}\hspace{0.85cm}\textbf{if} $\text{denom} \leq 0$ \textbf{then} \\
    {\small 12}\hspace{1.3cm}Raise safety violation and go to \textbf{step 26} \\
    {\small 13}\hspace{0.85cm}\textbf{else} \\
    {\small 14}\hspace{1.3cm}Compute safety metric: $\log\left(\frac{o^2}{o^2 - e_i^2}\right)$ \\
    {\small 15}\hspace{1.3cm}Update adaptive parameter: $\dot{\hat{\theta}}_i = -\delta_i \hat{\theta}_i + \gamma_i \left(\frac{e_i}{o^2 - e_i^2}\right)^2$ \\
    {\small 16}\hspace{1.3cm}Compute RAC control: $u_{s_i} = -\frac{1}{2}k_i e_i - \gamma_i \frac{e_i}{o^2 - e_i^2} \hat{\theta}_i$ \\
    {\small 17}\hspace{1.3cm}Set control command: $u_{c_i} = u_{s_i} = n_{p_i}$ \\
    {\small 18}\hspace{1.3cm}Return to \textbf{step 8} \\
    {\small 19}\hspace{0.85cm}\textbf{end if} \\
    {\small 20}\hspace{0.6cm}\textbf{else if} \\
    {\small 21}\hspace{0.85cm}Keep $\alpha_1 = 1$ and $\alpha_2 = 0$ \\
    {\small 22}\hspace{0.85cm}Pass $v_{\text{ref}_i}$ to LM-trained DNN policy to compute $u_{{\text{DNN}}_i}$ \\
    {\small 23}\hspace{0.85cm}Set control command: $u_{c_i} = u_{{DNN}_i} = n_{p_i}$ \\
    {\small 24}\hspace{0.85cm}Return to \textbf{step 4} \\
    {\small 25}\hspace{0.6cm}\textbf{end if} \\
    {\small 26}\hspace{0.6cm}\textbf{terminate operation}\\
    {\small 27}\hspace{0.15cm}\textbf{end for} \\
    \\
    \hline
  \end{tabular}
    \begin{tablenotes}
      \item[- $\zeta$ corresponds to the low-level safety layer, while $o$ represents the high-level safety layer.]
      \item[- Once $R \leq 0$ is detected, the system permanently switches to RAC for safety (latch-off behavior).]
      \item[- If RAC is active, the safety metric ensures operations remain within safe bounds; otherwise, the system shuts down.]
    \end{tablenotes}
  \end{threeparttable}
\end{table}

\begin{equation}
\begin{aligned}
\small
\label{54}
V_i =\frac{1}{2} \log \left(\frac{o_i^2}{o^2 - e^2_i}\right)+ \frac{1}{2} {A_i}\tilde{\theta}_i^2
\end{aligned}
\end{equation}
From \eqref{9} and \eqref{51}, we have
\begin{equation}
\begin{aligned}
\small
\label{55}
\dot{{V}}_i (t)=&\frac{-\dot{o} e_i^2}{o\left(o^2-e_i^2\right)}\\
&+ A_i \frac{{e}_i}{o^2 - e^2_i} [\alpha(t_s){u_{\text{DNN}}}_i(t_s) + \alpha_2(t_s) u_{s_i}(t)] \\
&+  \frac{{e}_i}{o^2 - e^2_i} [\tau^{-1} (F_i - v_i)- \dot{v}_{{ref}_i}] +A_i \tilde{\theta}_i \dot{\tilde{\theta}}_i
\end{aligned}
\end{equation}
From \eqref{39}, and at switching time $t_s$, we have
\begin{equation}
\begin{aligned}
\small
\label{56}
\dot{V}_i \leq & A_i\frac{e_i}{o^2-e_i^2} (-  \frac{1}{2} k_i e_i - \gamma_i  \frac{e_i}{o^2-e_i^2} \hat{\theta}_i)\\
&+|\frac{e_i}{o^2-e_i^2}|\hspace{0.1cm}|\hspace{0.1cm}A_i {u_{\text{DNN}}}_i(t_s) + \tau^{-1} (F_i - v_i)\\
& - \dot{v}_{{ref}_i}-\frac{\dot{o}\hspace{0.1cm} {e_i}}{o}\hspace{0.1cm}|-\delta_i {A_i} \tilde{\theta}^2_i+\gamma_i  {A_i} (\frac{\bar{e}_i}{o^2-e_i^2})^2  \tilde{\theta}_i \\
&- {A_i}\delta_i \theta^*_i \tilde{\theta}_i 
\end{aligned}
\end{equation}
Based on \textit{Assumption II.3}, we can assume there exist a positive
parameter $f^*_i$ such that
\begin{equation}
\begin{aligned}
\small
\label{57}
|\hspace{0.1cm}A_i {u_{\text{DNN}}}_i(t_s) + \tau^{-1} (F_i - v_i) - \dot{v}_{{ref}_i}-\frac{\dot{o}\hspace{0.1cm} {e_i}}{o}\hspace{0.1cm}| \leq f^*_i
\end{aligned}
\end{equation}
Based on \textit{Remark IV.1}, \eqref{56} and \eqref{57}, and knowing $\tilde{\theta}_i = \hat{\theta}_i - \theta^*_i$, we have
\begin{equation}
\begin{aligned}
\small
\label{58}
\dot{V}_i \leq & - \frac{1}{2} {A_i} k_i \frac{{e}^2_i}{o^2-e_i^2}  - {A_i} \gamma_i  (\frac{{e}_i}{o^2-e_i^2})^2 {\theta}^*_i   \\
&+  \epsilon_i (\frac{{e}_i}{o^2-e_i^2})^2 {f^*_i}^2 + \frac{1}{4 \epsilon_i}-\delta_i {A_i} \tilde{\theta}^2_i- {A_i}\delta_i \theta^*_i \tilde{\theta}_i 
\end{aligned}
\end{equation}
Let us define an unknown positive $\theta^*_i$ as
\begin{equation}
\begin{aligned}
\small
\label{59}
\theta^*_i = \frac{\epsilon_i {f^*_i}^2}{\gamma_i A_i}
\end{aligned}    
\end{equation}
So, from \eqref{58} and \eqref{59}, we have
\begin{equation}
\begin{aligned}
\small
\label{60}
\dot{V}_i \leq & - \frac{1}{2} {A_i} k_i \frac{{e}^2_i}{o^2-e_i^2} + \frac{1}{4 \epsilon_i}-\delta_i {A_i} \tilde{\theta}^2_i - {A_i}\delta_i \theta^*_i \tilde{\theta}_i 
\end{aligned}
\end{equation}
Based on \textit{Lemma IV.1}, we have
\begin{equation}
\begin{aligned}
\small
\label{61}
\dot{V}_i \leq & - \frac{1}{2} {A_i} k_i \log \left(\frac{o_i^2}{o^2-e_i^2}\right) + \frac{1}{4 \epsilon_i} -\delta_i {A_i} \tilde{\theta}^2_i\\
&- {A_i}\delta_i \theta^*_i \tilde{\theta}_i 
\end{aligned}
\end{equation}
Evidently,
\begin{equation}
\begin{aligned}
\small
\label{62}
\dot{V}_i \leq & - \frac{1}{2} {A_i} k_i  \log \left(\frac{o_i^2}{o^2-e_i^2}\right)  + \frac{1}{4 \epsilon_i} -\frac{1}{2}\delta_i {A_i} \tilde{\theta}^2_i \\
& - \frac{1}{2} {A_i} {\delta}_i (\hat{\theta}_i - \theta^*_i)^2 - {A_i} {\delta}_i \theta^*_i (\hat{\theta}_i - \theta^*_i)
\end{aligned}
\end{equation}
Similarly, to \eqref{48}, we obtain
\begin{equation}
\begin{aligned}
\small
\label{63}
\dot{V}_i \leq & - \frac{1}{2} {A_i} k_i  \log \left(\frac{o_i^2}{o^2-e_i^2}\right)  + \frac{1}{4 \epsilon_i} -\frac{1}{2}\delta_i {A_i} \tilde{\theta}^2_i \\
& + \frac{1}{2} A_i {\delta_i \theta^*_i}^2 
\end{aligned}
\end{equation}
Finally, we have
\begin{equation}
\begin{aligned}
\small
\label{64}
\dot{V}_i \leq & - \mu_i \mathcal{V}_i + \ell_i 
\end{aligned}    
\end{equation}
where $\mu_i=\min[A_i k_i, \delta_i]$ and $\ell_i = \frac{1}{4 \epsilon_i} + \frac{1}{2} A_i {\delta_i \theta^*_i}^2$. Thus, based on the resulting Eq. \eqref{64} and \cite{shahna2025robudfdsfst, shahasaasca2024asdacvrobustness, shahna2024exponential, corless1993bounded, shahna2024integrating, wu2020robust}, the proposed synthesis of the DNN and RAC policy for the SSWMR system, as provided in Eq. \eqref{9}, is uniformly exponentially stable even while considering switching time.

Figure \ref{i9sasddadado_synthes} illustrates the hierarchical structure of the proposed safe control architecture for the studied SSWMR. The framework integrates two safety layers—a low-level safety layer and a high-level safety layer—to ensure reliable operation under both nominal and perturbed conditions. Under nominal conditions, the LM-trained DNN control policy acts as the main controller, providing accurate control commands based on the reference tracking error. Its output is continuously monitored by the latched safety layer, which detects performance deviations using real-time sensor feedback. If predefined control metrics are violated, the low-level safety layer is triggered. If the disturbance is compensable, the RAC policy (located in the high-level safety layer) is activated to take control and enhance robustness. However, if the disturbance exceeds the compensatory capacity of the RAC, the high-level safety layer—based on a persistent safety violation—triggers a shutdown to prevent system or environmental damage. This leads to a safe termination, as indicated in red, maintaining safety-critical guarantees in both normal and adverse operating scenarios. 

\section{Experimental Validity}
\label{expmsdnaf}

\begin{figure*}[h!]
\hspace*{-0.0cm} 
\centering
\includegraphics[width=0.7\textwidth, height=8cm]{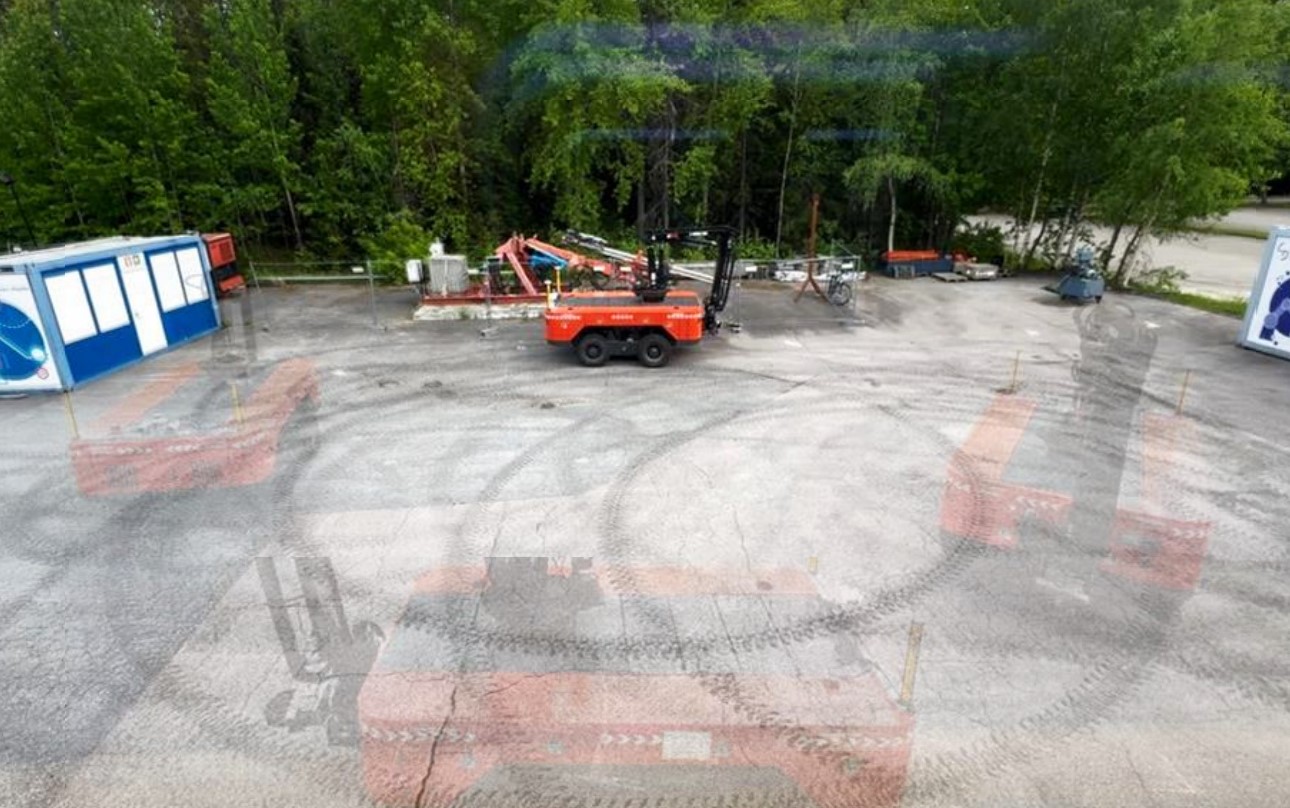}
\caption{Circular trajectory (6 m radius) tracked by a 6000-kg SSWMR using the LM-trained DNN control policy (Exp. 1) and the safe RAC policy (Exp. 2) under one safety layer.}
\label{circcccular}
\end{figure*}

In this section, we validate the proposed control policies through implementation on a $6,000$ kg PMSM-powered heavy-duty SSWMR, referred to as the multi-purpose deployer (MPD). Each side of the robot features bogie-mounted wheels that maintain ground contact during traversal. The mechanical structure and sensor setup of the MPD is the same as presented in Figs. \ref{fig111sadsad521f} and \ref{aFfa}. The electrical system comprises a battery-powered inverter that precisely regulates the speed and torque of each PMSM. The hydraulic system transfers power from the pumps to the hydraulic motors. Each hydraulic motor is equipped with an odometry sensor, and the velocities from the two wheels on each side are averaged to determine the measured velocity of that side at each time step. This process encompasses energy conversion, high-torque motor operation, high-ratio gear transmission, and ultimately, wheel actuation. An embedded Beckhoff PC handles sensor data acquisition via the EtherCAT protocol. For the purposes of control, we assume that all modeling details of the MPD are unknown. The system has access only to speed measurements from each side and can control the PMSM through its rpm input.

\subsection{Two LM Training DNNs for the MPD}
\label{Exp1}
For each side of the MPD, based on \textbf{Algorithm 1}, we first collected a dataset of $1{,}000{,}000$ synchronized input–output pairs by slightly increasing the control inputs $n_{p_R}$ and $n_{p_L}$ (PMSM rpm) and recording the corresponding actual linear velocities $v_R$ and $v_L$ from the installed sensor during various operational scenarios. These recorded data served as the foundation for model training. On the input side of the training models, we logged the linear velocity of each track using Danfoss EMD speed sensors mounted on the in-wheel hydraulic motors. For the outputs, we recorded the control signal to each side’s PMSM-driven pump actuator—namely the pump-shaft rotational speed in rpm. All measurements were timestamped at a fixed sampling interval of 0.001 s (1 kHz), handled in real time by a Beckhoff IPC CX2030. Raw velocity and rpm streams were then aligned, cleaned of obvious spurious spikes, and split into training, validation, and test subsets before normalization.

Second, starting from the raw $1 \times \mathrm{1,000,000}$ data vectors $\mathrm{v}_i$ (sensor readings) and $\mathrm{n}_{p_i}$ (control signals), we first randomly partitioned the sample indices into $70 \%$ training, $15 \%$ validation, and $15 \%$ test sets using MATLAB's \textit{dividerand} function (see \textit{Appendix A}). Then, the corresponding input/output train, input/output validation, and input/output test subsets were extracted, still in their original ("raw") units. Next, we normalized only the input-output train to the $[-1,+1]$ range with MATLAB's \textit{mapminmax} function, saving the scaling parameters. We applied these same parameters to map validation and test inputs into the identical normalized space.

Third, we built LM training DNNs in MATLAB with five hidden layers of sizes $[30, 25, 15, 10, 5]$, using the default \textit{tan-sigmoid} activations in the hidden layers and a linear output neuron. All built-in input/output processing functions were disabled because we fed the network pre-scaled data directly.
Training employed the scaled conjugate gradient (\textit{trainscg}) algorithm, which offers fast second-order-like convergence without storing a full Hessian.
We set the performance goal of mean-squared error to $1 \times 10^{-3}$, a minimum gradient threshold of $1 \times 10^{-4}$, and a hard cap of 200 epochs. Validation checks were enabled via a 70/15/15 train/validation/test split; thus, the network also monitored validation MSE for early stopping (up to the default 6-failure limit) to guard against overfitting.

\begin{figure}[h!]
\hspace*{-0.0cm} 
\centering
\scalebox{1.0}{\includegraphics[trim={0cm 0.0cm 0.0cm 0cm},clip,width=\columnwidth]{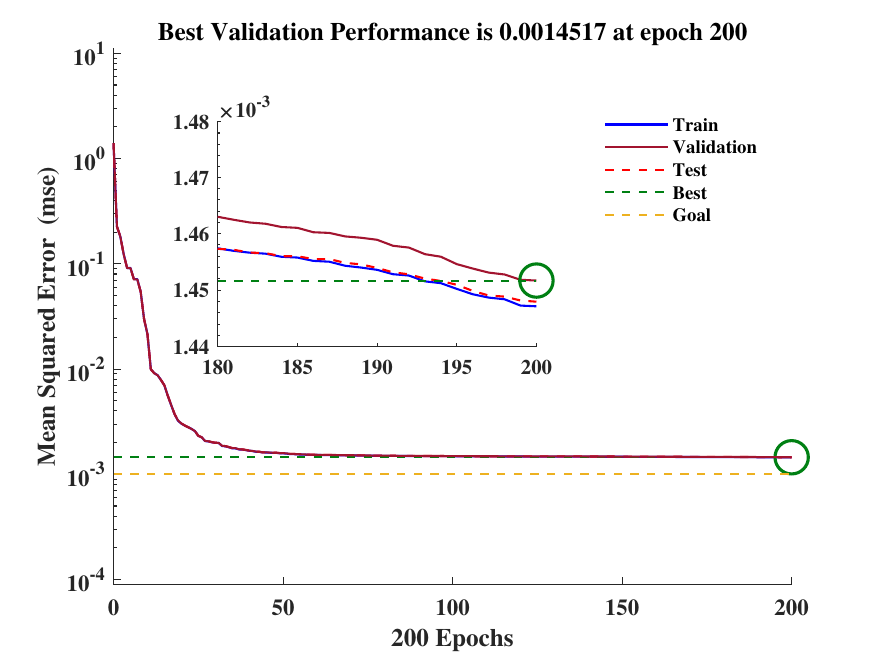}}
\caption{Performance of LM training DNN for right side data.}
\label{performance_right}
\end{figure}

\begin{figure}[h!]
\hspace*{-0.0cm} 
\centering
\scalebox{1.0}{\includegraphics[trim={0cm 0.0cm 0.0cm 0cm},clip,width=\columnwidth]{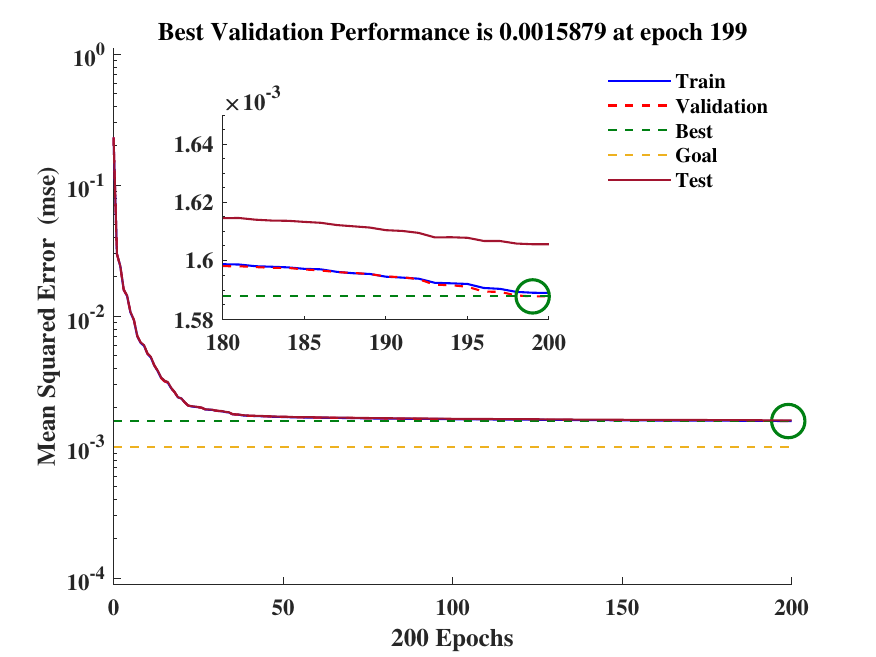}}
\caption{Performance of LM training DNN for left side data.}
\label{performance_left}
\end{figure}

\begin{figure}[h!]
\hspace*{-0.0cm} 
\centering
\scalebox{1.0}{\includegraphics[trim={0cm 0.0cm 0.0cm 0cm},clip,width=\columnwidth]{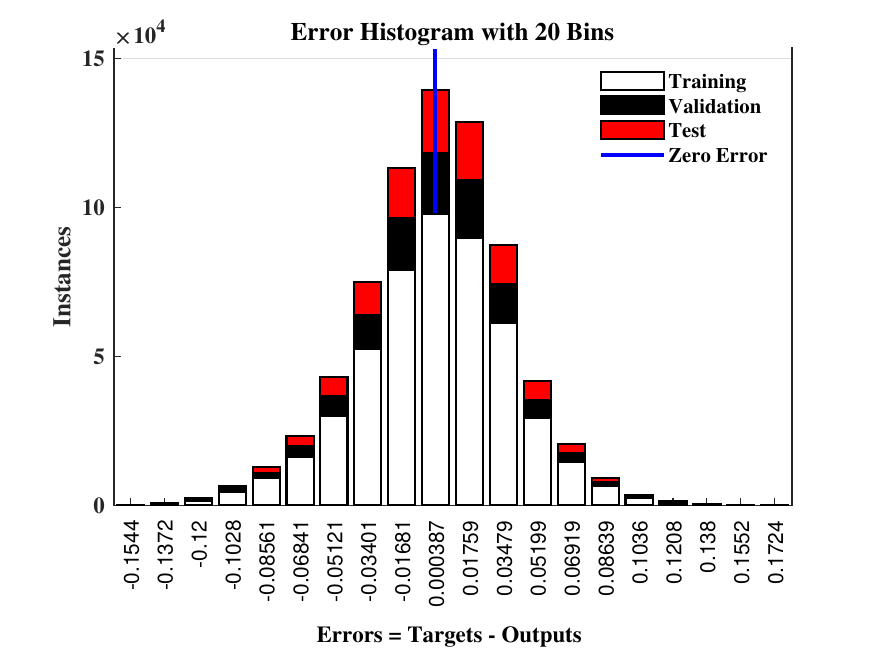}}
\caption{History of LM training DNN for right-side data.}
\label{history_right}
\end{figure}

\begin{figure}[h!]
\hspace*{-0.0cm} 
\centering
\scalebox{1.0}{\includegraphics[trim={0cm 0.0cm 0.0cm 0cm},clip,width=\columnwidth]{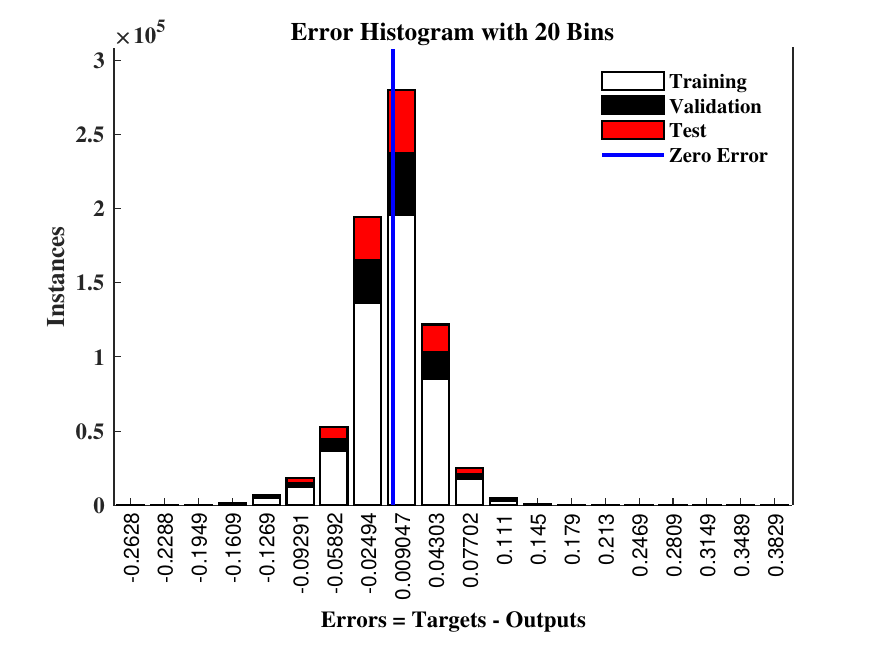}}
\caption{History of LM training DNN for left-side data.}
\label{history_left}
\end{figure}

\begin{figure}[h!]
\hspace*{-0.0cm} 
\centering
\scalebox{1.0}{\includegraphics[trim={0cm 0.0cm 0.0cm 0cm},clip,width=\columnwidth]{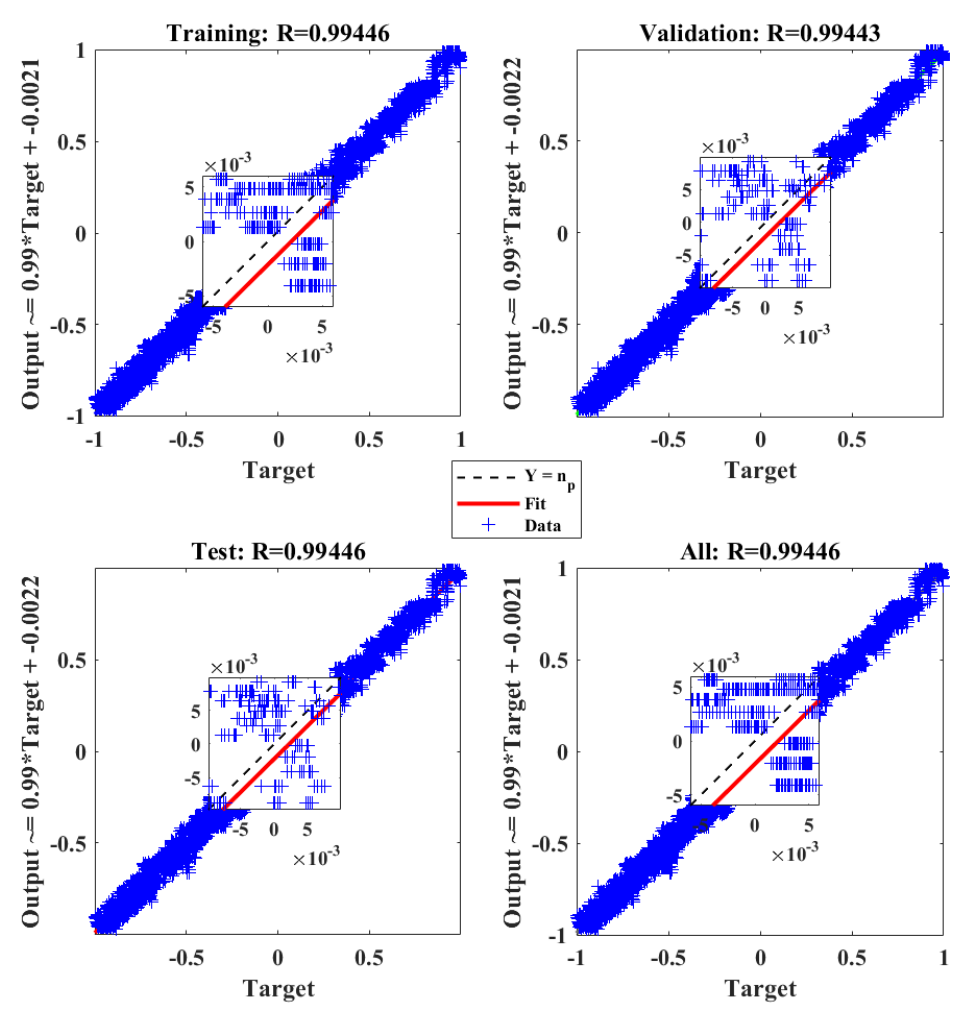}}
\caption{Regression of LM training DNN for right-side data.}
\label{regression_right}
\end{figure}

\begin{figure}[h!]
\hspace*{-0.0cm} 
\centering
\scalebox{1.0}{\includegraphics[trim={0cm 0.0cm 0.0cm 0cm},clip,width=\columnwidth]{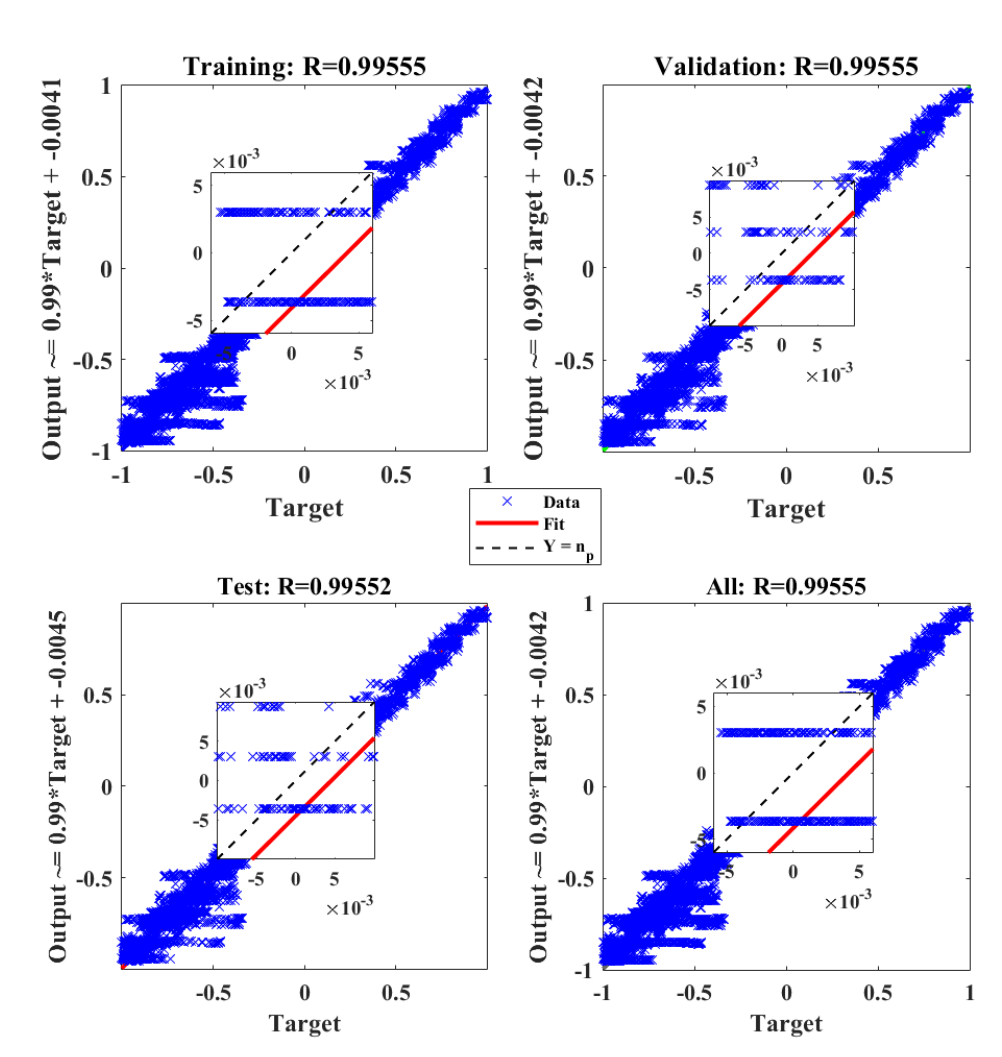}}
\caption{Regression of LM training DNN for left-side data.}
\label{regression_left}
\end{figure}

\begin{figure}[h!]
\hspace*{-0.0cm} 
\centering
\scalebox{1.0}{\includegraphics[trim={0cm 0.0cm 0.0cm 0cm},clip,width=\columnwidth]{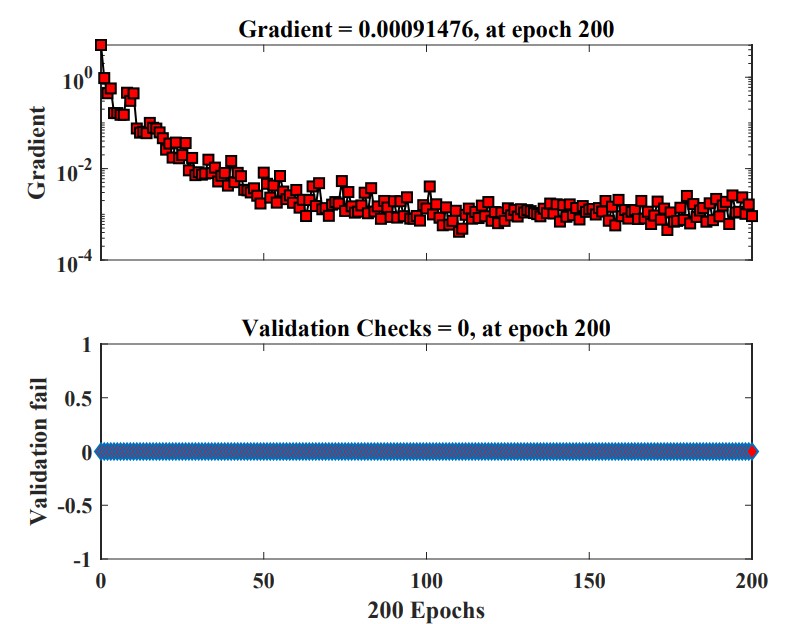}}
\caption{Training state of LM training DNN for right-side data.}
\label{steady_right}
\end{figure}

\begin{figure}[h!]
\hspace*{-0.0cm} 
\centering
\scalebox{1.0}{\includegraphics[trim={0cm 0.0cm 0.0cm 0cm},clip,width=\columnwidth]{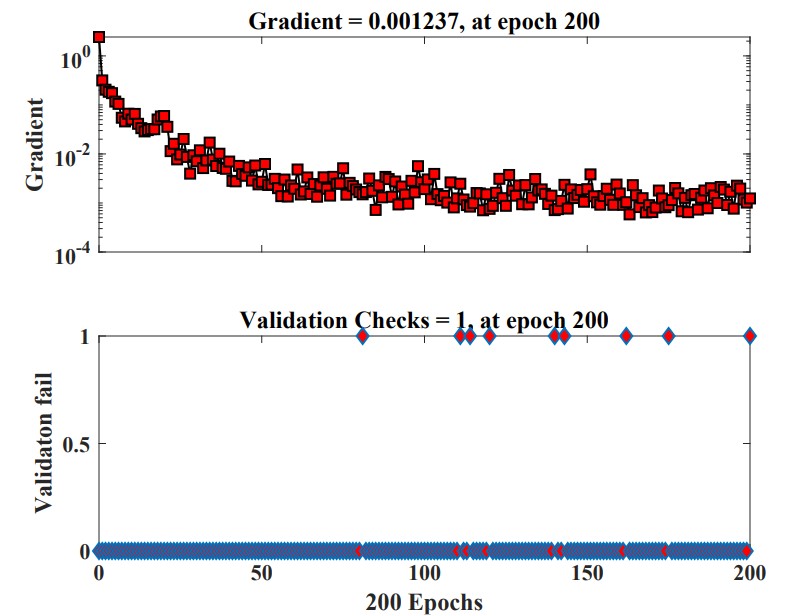}}
\caption{Training state of LM training DNN for left-side data.}
\label{steady_left}
\end{figure}

Figures \ref{performance_right}-\ref{steady_left} indicate the relevant plots after two training processes within the simulation environment. Figs. \ref{performance_right} and \ref{performance_left} show that over 200 epochs the network's training, validation, and test MSE all drop rapidly in the first few dozen iterations and then gently taper off to around $1.5 \times 10^{-3}$, with the best validation error ( $\approx 0.00145$ ) occurring at epoch 200; the validation and test curves remain closely tracking the training curve without any upward turn, indicating steady learning and no overfitting, and since two models have not yet plateaued sharply or begun to diverge, it's both well-trained and stable.

Figures \ref{history_right} and \ref{history_left} display the error histograms for the right and left sides of WMR training, respectively. Both distributions are tightly centered and approximately symmetric around zero, indicating balanced model behavior with minimal bias. In Fig. \ref{history_right}, the majority of target–output residuals fall within the $±0.02$ range, peaking in the bin centered at zero and aligning with the blue zero-error line. Similarly, Fig. \ref{history_left} shows a sharply peaked distribution with an even narrower error spread, mostly within $±0.01$, and a peak slightly offset from zero but still closely aligned with the zero-error reference.

In both cases, training (white), validation (black), and test (red) errors are stacked consistently across bins, suggesting that the models generalize well with low variance across all data splits. While the right-side model has very few errors exceeding $±0.05$, the left-side model shows even fewer outliers, with almost none beyond $±0.1$. Overall, both models demonstrate strong accuracy, low error magnitudes, and consistent performance across data partitions, with the left-side model showing marginally tighter error concentration.

Figures \ref{regression_right} and \ref{regression_left} show regression plots for the right and left sides of WMR training, respectively. In both figures, the predicted outputs align closely with the ideal 45° reference line (dashed), with all subsets—training, validation, and test—exhibiting strong linear relationships between predicted and target values. The correlation coefficients are exceptionally high in both cases, with $R \approx 0.9945$ for the right side and slightly higher at $R \approx 0.9955$ for the left, indicating equally strong predictive performance across models. The fitted lines in both figures have slopes near 0.99 and small intercepts near zero, suggesting minimal underestimation bias. Insets further highlight the tight concentration of residuals around the ideal line. Notably, the left-side model (Fig. \ref{regression_left}) achieves a marginally better fit, evidenced by its slightly higher R-values and more tightly clustered residuals. Nevertheless, both models demonstrate excellent accuracy and generalization across training, validation, and test sets, with no indication of overfitting or systematic bias.

Figures \ref{steady_right} and \ref{steady_left} illustrate the training dynamics for the right and left sides of the WMR model, respectively, over 200 epochs. In both cases, the top plots show a steady and significant decline in gradient magnitude, converging to the order of $10^{-3}$ by the end of training. This indicates effective optimization with stable gradient descent behavior. Specifically, the final gradient is approximately $0.00091$ for the right side and $0.00124$ for the left—both sufficiently small to indicate convergence.
The lower subplots display validation check indicators across epochs. For the right-side model (Fig. \ref{steady_right}), no validation failures occur throughout training, suggesting that the model maintained consistent generalization without any regressions on the validation set. The left-side model (Fig. \ref{steady_left}), however, exhibits a single validation failure near epoch 100, but this isolated event did not result in early stopping, as the model continued training to epoch 200. The minimal number of validation checks (zero or one) in both models supports the observation of smooth and stable training with no significant overfitting.

\subsection{Experiment 1: Safe LM-trained DNN-based Control Policy for Circular Path}
\label{expdnn}
\begin{figure}[h!]
\hspace*{-0.0cm} 
\centering
\scalebox{1.0}{\includegraphics[trim={0cm 0.0cm 0.0cm 0cm},clip,width=\columnwidth]{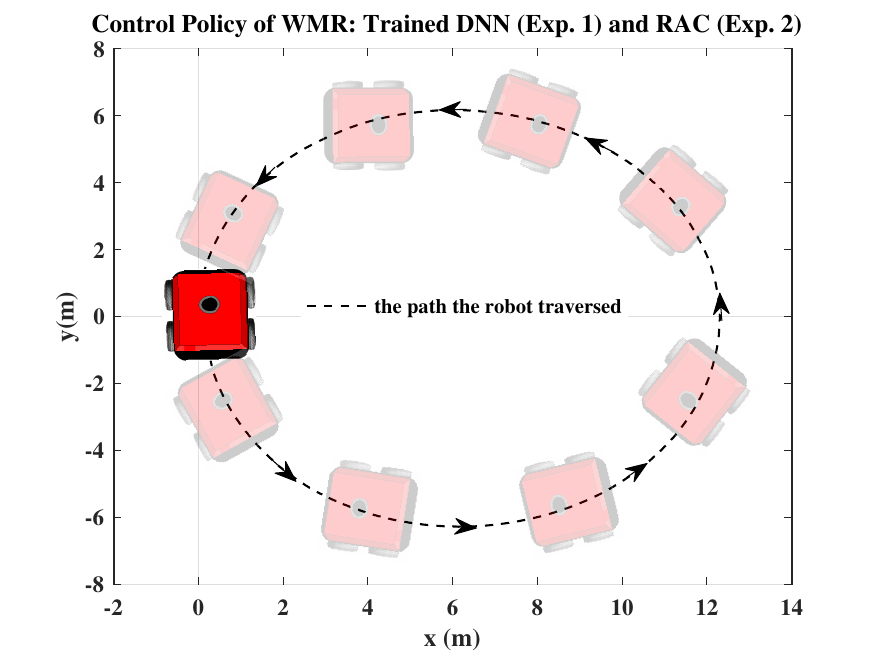}}
\caption{Operating task in Experiments 1 and 2.}
\label{slam_sicrlc}
\end{figure}

To evaluate the practical performance, the two trained DNN-based control policies for the right and left sides of the MPD in Section \ref{Exp1} were deployed to follow a circular path with an approximate radius of $6$ m (see Fig. \ref{circcccular}). The velocity reference for each side was generated using the method described in Ref. \cite{shahna2025lidar}.
Figure \ref{slam_sicrlc} shows the demanding trajectory of the WMR operating under the trained DNN-based control policy. The robot starts at origin $(0,0)$ and follows a smooth, continuous path, completing a loop of approximately $12$ m in diameter. The dashed black line indicates the path the robot should traverse. 
Two side-specific trained DNN control policies received the commanded wheel velocities directly and were responsible for generating the corresponding PMSM control inputs required to track those velocities. 

\begin{figure}[h!]
\hspace*{-0.0cm} 
\centering
\scalebox{1.0}{\includegraphics[trim={0cm 0.0cm 0.0cm 0cm},clip,width=\columnwidth]{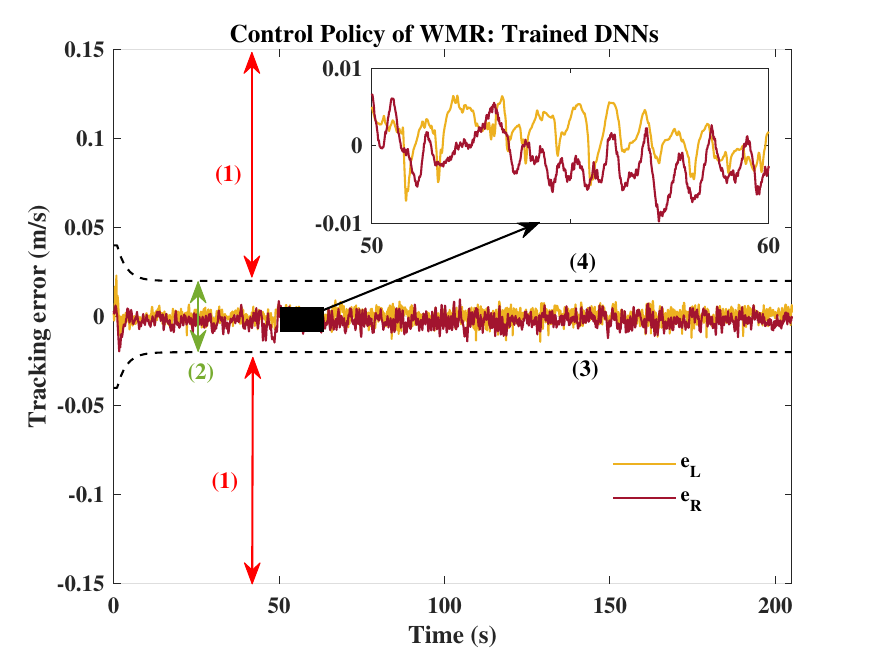}}
\caption{Exp.1. Tracking error under LM-trained DNN control policies. \textcolor{red}{(1)}: Forbidden performance, \textcolor{matlabgreen}{(2)}: permissible performance, (3): permissible limit for the control policy $\zeta = -0.02 e^{-0.35t} - 0.02$, (4): permissible limit for the control policy $\zeta = 0.02 e^{-0.35t} + 0.02$.}
\label{DNN_error}
\end{figure}

To ensure safe operation (see \textbf{Algorithm 2}), the safety layer $\zeta$ was employed to guarantee that the system operates within a predefined safe region. Following this, Fig. \ref{DNN_error} displays the tracking error of the MPD using the two-side trained DNN policy, under performance boundaries defined. The tracking errors for the left and right sides, $e_L$ (yellow) and $e_R$ (maroon), are plotted over $200$ s. The graph includes time-varying performance bounds: the lower limit ${\zeta}=$ $-0.02 e^{-0.1 t}-0.02$ and upper limit ${\zeta}=0.02 e^{-0.1 t}+0.02$, shown as dashed black curves (labels 3 and 4). The central region between these bounds represents permissible performance, while the areas outside are marked as forbidden. During the circular operation shown in Fig. \ref{slam_sicrlc}, the errors remained within the safe region, with no violations, validating the significant accuracy of the trained model. An inset zoom-in between 50 and 60 s highlights the fine-scale behavior, showing perfect tracking control for such a complex and heavy WMR.

\begin{figure}[h!]
\hspace*{-0.0cm} 
\centering
\scalebox{1.0}{\includegraphics[trim={0cm 0.0cm 0.0cm 0cm},clip,width=\columnwidth]{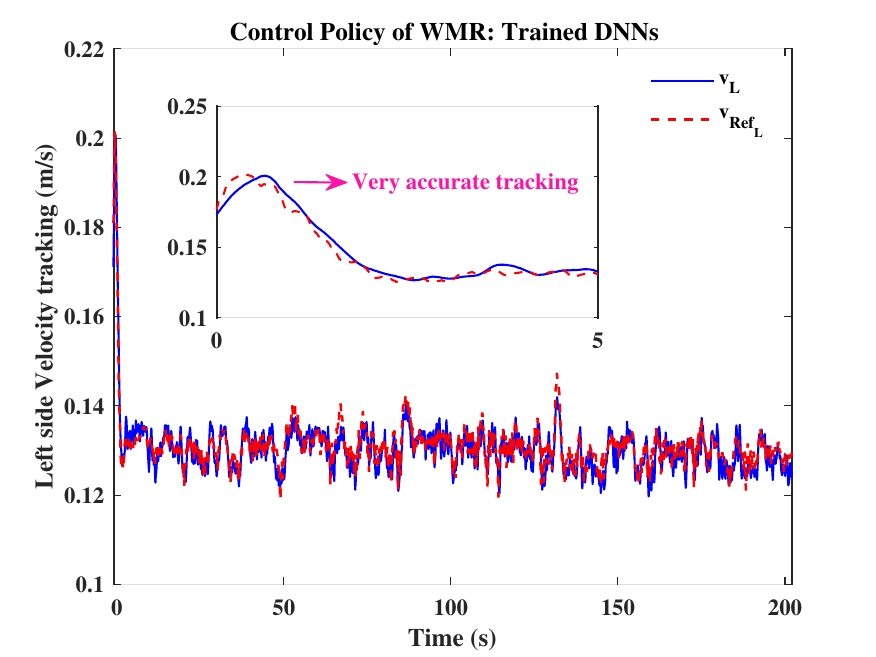}}
\caption{Exp.1. Tracking in the left side under LM-trained DNN control policy}
\label{DNN_velleft}
\end{figure}

\begin{figure}[h!]
\hspace*{-0.0cm} 
\centering
\scalebox{1.0}{\includegraphics[trim={0cm 0.0cm 0.0cm 0cm},clip,width=\columnwidth]{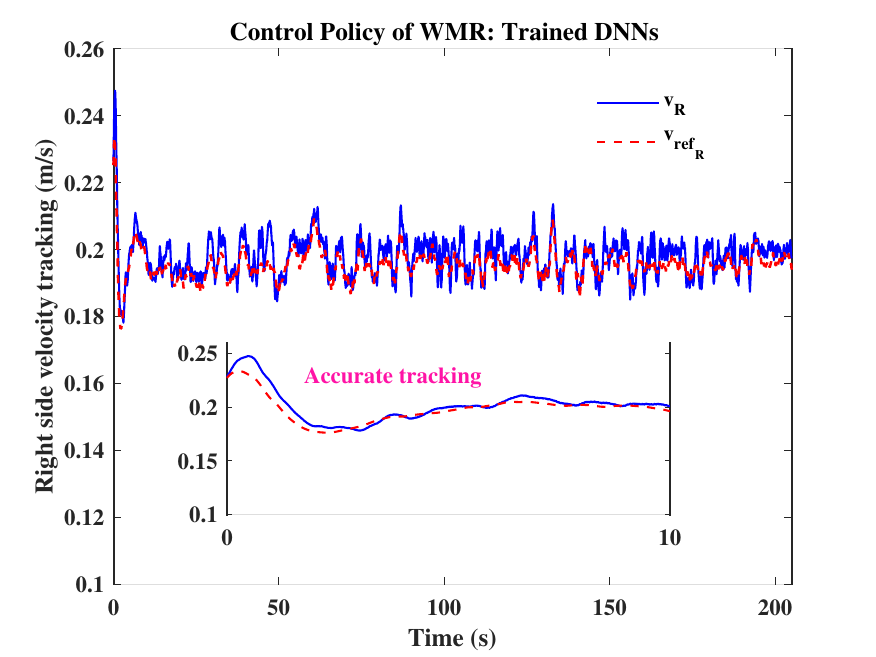}}
\caption{Exp.1. Tracking in the right side under LM-trained DNN control policy}
\label{DNN_velright}
\end{figure}

Figures \ref{DNN_velleft} and \ref{DNN_velright} illustrate the left- and right-side velocity tracking performance of the WMR under trained DNNs. In both cases, the actual wheel velocities $\left(v_L\right.$ and $v_R$, solid blue lines) are compared to their respective reference signals ( $v_{\operatorname{Ref}_L}$ and $v_{\operatorname{Ref}_R}$, dashed red lines) over a 200-s period. Despite the black-box nature of the control policy, both side controllers achieve excellent tracking performance. After an initial transient phase, the velocities stabilize and closely follow their references throughout the entire experiment. The insets in each figure, highlighting the first few seconds, reveal high-fidelity tracking with minimal deviation, particularly in the left-side controller, which exhibits slightly smoother convergence. The right-side controller also maintains strong alignment, albeit with slightly more oscillatory behavior due to the nature of the curve direction in the task. Overall, these results demonstrate the ability of the DNN-based control policy to generalize effectively to both sides of the WMR, achieving accurate and stable velocity tracking under nominal conditions (absence of severe disturbances or faults).

\begin{figure}[h!]
\hspace*{-0.0cm} 
\centering
\scalebox{1.0}{\includegraphics[trim={0cm 0.0cm 0.0cm 0cm},clip,width=\columnwidth]{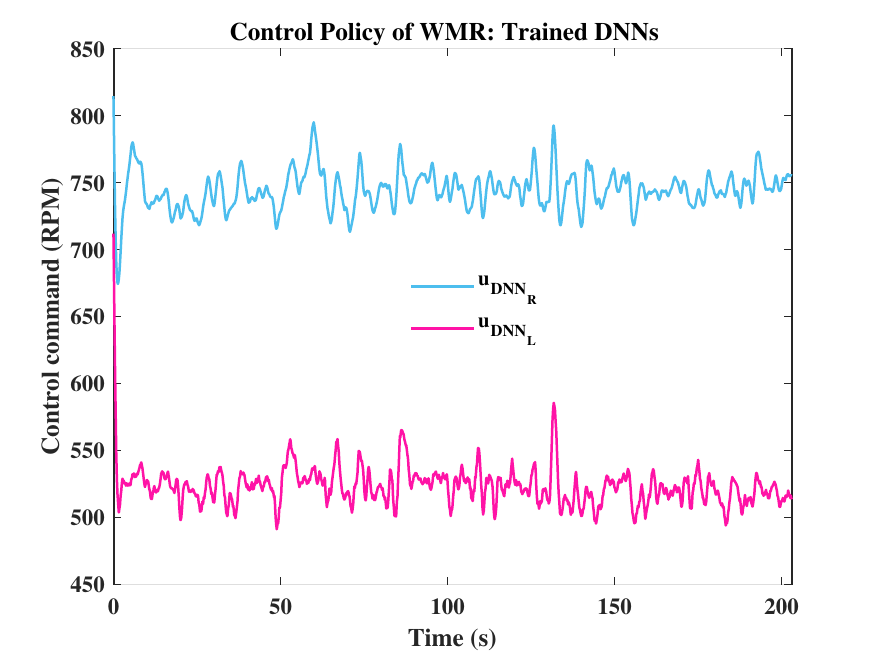}}
\caption{Exp.1. Control commands (PMSM output) under LM-trained DNN control policies.}
\label{dnn_contcom}
\end{figure}

Figure \ref{dnn_contcom} displays the control input signal (PMSM output) generated by the LM-trained DNN control policies for the right and left wheels of the WMR. The plot shows $u_{\text{DNN}_R}$ (right wheel, light blue) and $u_{\text{DNN}_L}$ (left wheel, magenta) in rpm over a 200-s time span. Both signals exhibit stable, bounded behavior with moderate fluctuations around their respective mean values of approximately 730 rpm for the right wheel and 520 rpm for the left (greater effort from the right side helps the robot steer along leftward trajectories). This indicates the required dynamics while avoiding abrupt or erratic changes.

\subsection{Experiment 2: Safe RAC Policy for Circular Path}
\label{exppraac}
\begin{figure}[h!]
\hspace*{-0.0cm} 
\centering
\scalebox{1.0}{\includegraphics[trim={0cm 0.0cm 0.0cm 0cm},clip,width=\columnwidth]{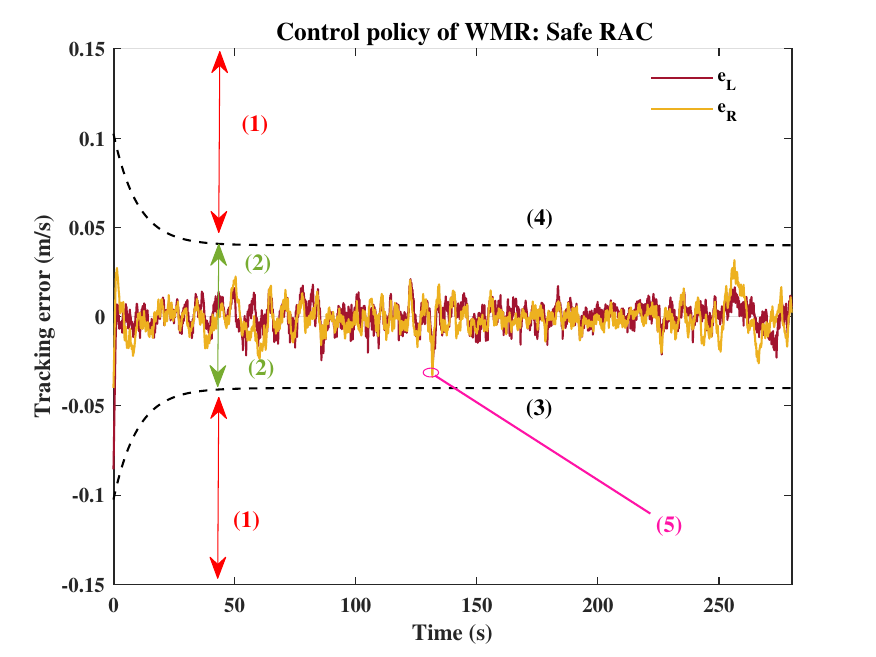}}
\caption{Exp. 2. Tracking error under safe RAC policies. \textcolor{red}{(1)}: Forbidden performance, \textcolor{matlabgreen}{(2)}: permissible performance, (3): permissible limit for the control policy ${o} = -0.06 e^{-0.1t} - 0.04$, (4): permissible limit for the control policy ${o} = 0.06 e^{-0.1t} + 0.06$, \textcolor{darkpink}{(5)}: behavior of the tracking-error response when the external disturbance occurred at 130
s.}
\label{rac_error}
\end{figure}

Similar to Section \ref{expdnn}, to evaluate practical performance, the proposed safe RAC scheme was also deployed on the experimental WMR platform to follow the same circular path as that shown in Figs. \ref{circcccular} and \ref{slam_sicrlc}. The plot in Fig. \ref{rac_error} presents the left and right wheel tracking errors, $e_L$ and $e_R$, in maroon and yellow, respectively. The chart is overlaid with performance bounds: the inner dashed lines represent the time-varying permissible limits defined by exponential decay functions ${o}=$ $-0.06 e^{-0.1 t}-0.04$ and ${o}=0.06 e^{-0.1 t}+0.04$, which gradually tighten over time. 

\begin{figure}[h!]
\hspace*{-0.0cm} 
\centering
\scalebox{1.0}{\includegraphics[trim={0cm 0.0cm 0.0cm 0cm},clip,width=\columnwidth]{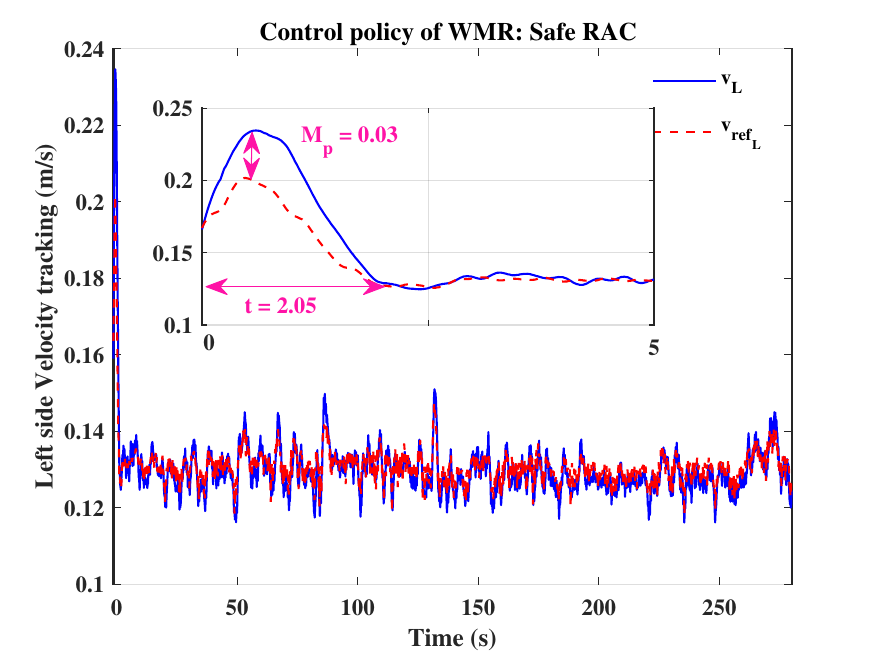}}
\caption{Exp.2. Tracking in the left side under safe RAC policy}
\label{lef_racvel}
\end{figure}

\begin{figure}[h!]
\hspace*{-0.0cm} 
\centering
\scalebox{1.0}{\includegraphics[trim={0cm 0.0cm 0.0cm 0cm},clip,width=\columnwidth]{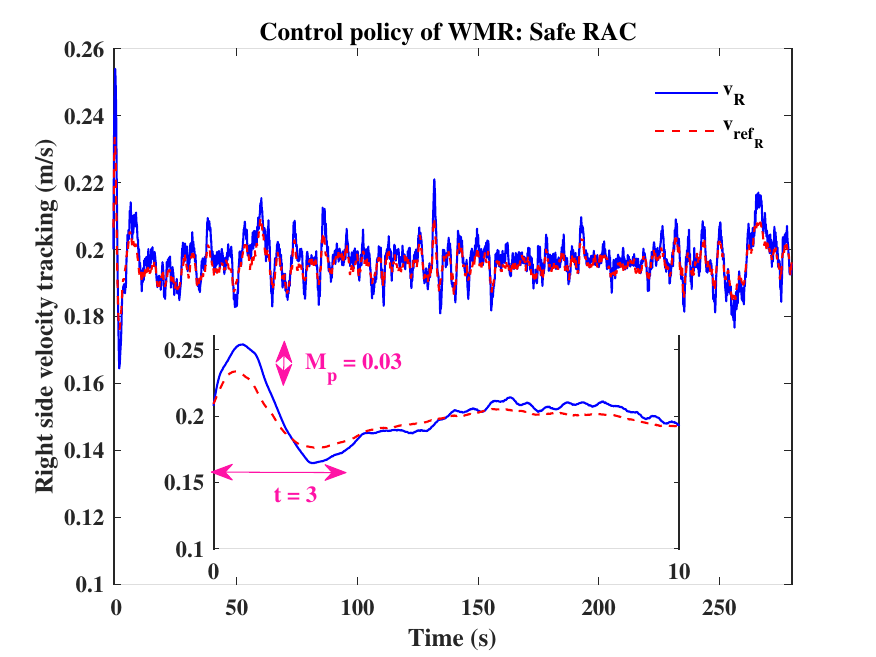}}
\caption{Exp.2. Tracking in the left side under safe RAC policy}
\label{right_racvel}
\end{figure}

The areas outside these bounds are labeled as forbidden performance zones, while the region between the bounds is marked as permissible. At $t=130$ s, two small external disturbances were introduced by increasing the control input by $8 \%$ relative to the commanded values. Despite these disturbances, both tracking error curves remained well within the defined performance bounds throughout the experiment. This demonstrates that the safe RAC policy effectively compensated for the disturbances while adhering to performance constraints and maintaining tracking errors within the safety margins. As mentioned earlier, although the adaptive laws in the RAC scheme can improve robustness against uncertainties and external disturbances, it comes at the cost of reduced control accuracy compared to the DNN-based policy shown in Fig. \ref{DNN_error}, highlighting a trade-off between robustness and responsiveness. This is evident in Fig. \ref{rac_error}, where the tracking error under the safe RAC scheme is higher than that of the trained DNN policies. As a result, the performance bounds for the RAC were set wider than those used for the DNN approach. A similar conclusion is obtained in Figs. \ref{lef_racvel} and \ref{right_racvel}.
These figures illustrate the left and right wheel velocity tracking performance of the WMR platform under the proposed safe RAC scheme.

\begin{figure}[h!]
\hspace*{-0.0cm} 
\centering
\scalebox{1.0}{\includegraphics[trim={0cm 0.0cm 0.0cm 0cm},clip,width=\columnwidth]{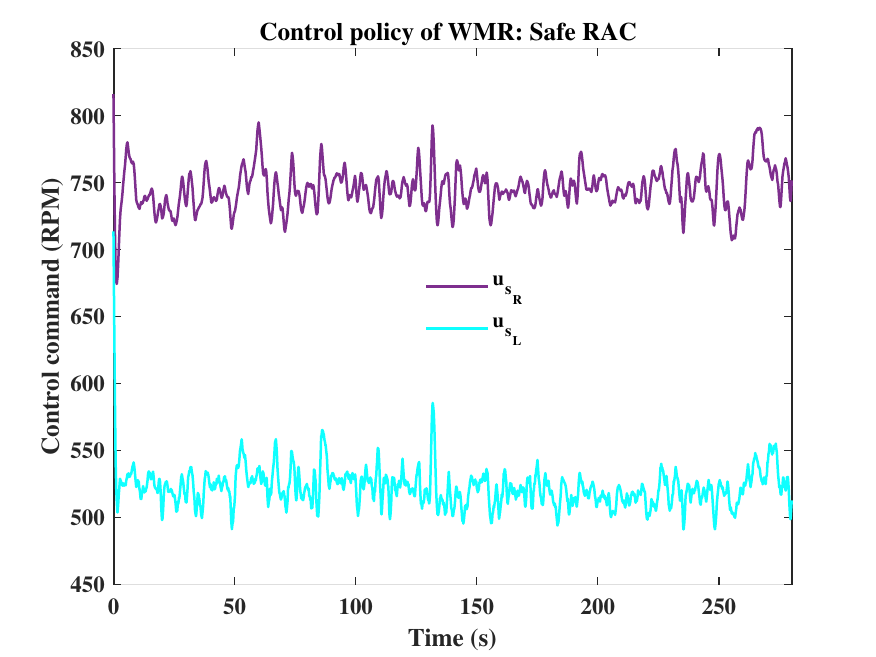}}
\caption{Exp.2. Control commands (PMSM output) under safe RAC policies.}
\label{control_rac_comand}
\end{figure}

The two plots exhibit similar trends. Both figures exhibit an initial transient phase characterized by slight overshooting and a short settling time. Specifically, Fig. \ref{lef_racvel} shows a left wheel overshoot ( $M_p$ ) of approximately $0.03 \mathrm{~m} / \mathrm{s}$ and a settling time of approximately 2.05 s, whereas Fig. \ref{right_racvel} displays the same overshoot magnitude for the right wheel, with a slightly longer settling time of approximately 3 s. During the steady-state phase (after the transient), both actual velocities closely track their reference values, with minor oscillations and fluctuations likely due to adaptive laws to compensate for unmodeled dynamics. These two figures also reveal that, although the tracking performance is acceptable, the control characteristics-particularly overshooting and transient response-are slightly weaker compared to the trained DNN-based controllers, as shown in Figs. \ref{DNN_velleft} and \ref{DNN_velright}.
Figure \ref{control_rac_comand} depicts the control commands generated by the safe RAC controller for the left and right wheels of the WMR. Despite some noise and fluctuations, the control signals are smooth and stable overall, indicating effective regulation by the safe RAC policy. 

\begin{figure}[h!]
\hspace*{-0.0cm} 
\centering
\scalebox{1.0}{\includegraphics[trim={0cm 0.0cm 0.0cm 0cm},clip,width=\columnwidth]{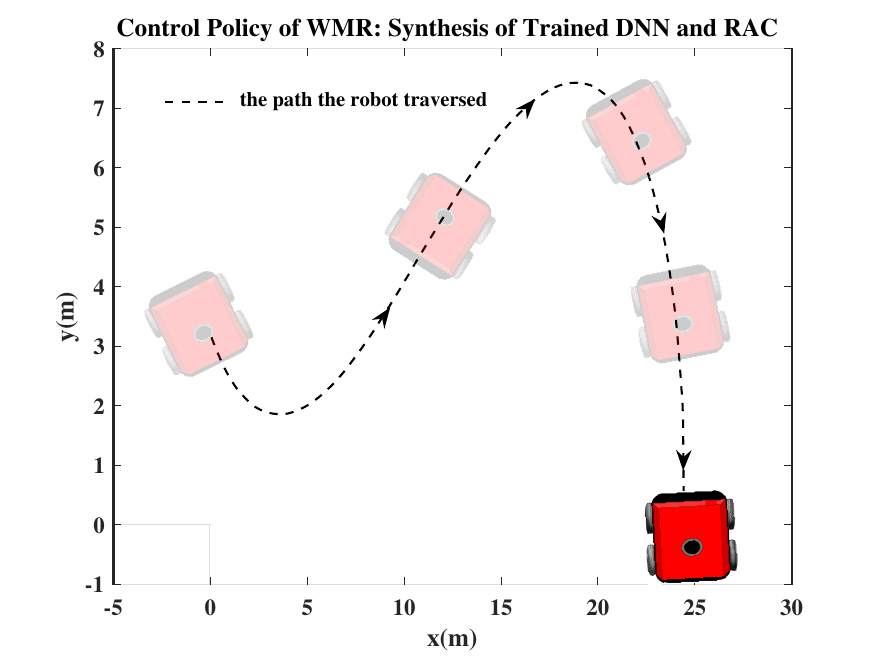}}
\caption{Operating task in Experiment 3.}
\label{salam_s_shap}
\end{figure}

\begin{figure*}[h!]
\hspace*{-0.0cm} 
\centering
\includegraphics[width=0.7\textwidth, height=8cm]{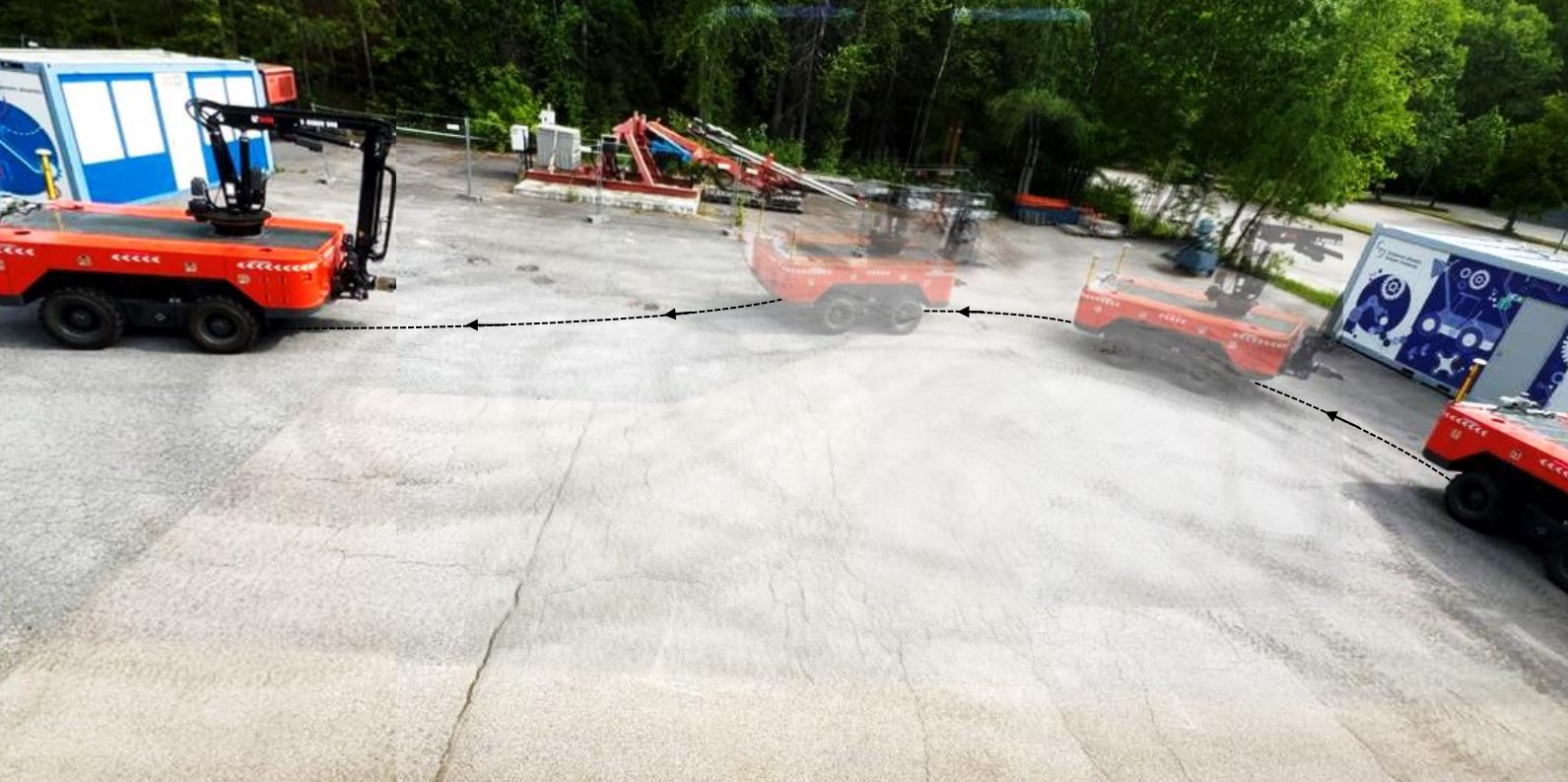}
\caption{Exp. 3. S-shape trajectory tracked by the SSWMR using the synthesis of the LM-trained DNN and the safe RAC policy under two safety layers.}
\label{control.s}
\end{figure*}

\subsection{Experiment 3: Synthesis of Trained DNN Control and Safe RAC Policy for S-shaped Path Tracking under Consistent Disturbance}
\begin{figure}[h!]
\hspace*{-0.0cm} 
\centering
\scalebox{1.0}{\includegraphics[trim={0cm 0.0cm 0.0cm 0cm},clip,width=\columnwidth]{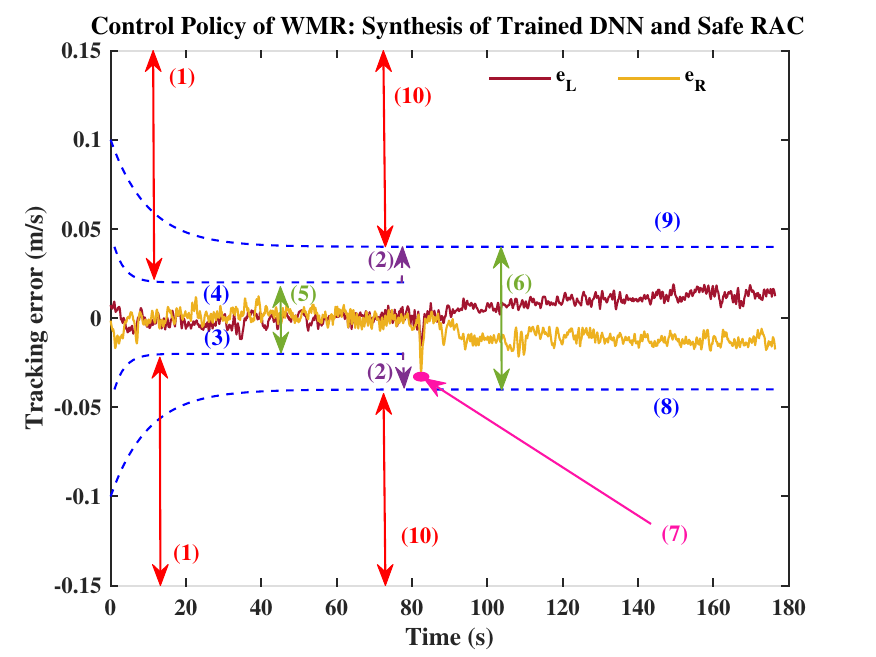}}
\caption{Exp. 3. Tracking error under the synthesis of LM-trained DNN and safe RAC policy. \textcolor{red}{(1)}: Forbidden performance for the DNN control policy, \textcolor{matlabpurple}{(2)}: switching control activation because of external disturbance (exceeding the low-level safety layer), \textcolor{blue}{(3)}: permissible lower limit for the DNN policy ${\zeta}= -0.02 e^{-0.35t} - 0.02$,  \textcolor{blue}{(4)}: permissible upper limit for the DNN policy ${\zeta}= 0.02 e^{-0.35t} + 0.02$, \textcolor{matlabgreen}{(5)}: permissible performance for the low-level safety layer, \textcolor{matlabgreen}{(6)}: permissible performance for the high-level safety layer, \textcolor{darkpink}{(7)}: behavior of the tracking‐error response when the external disturbance occurred at $82$ s,  \textcolor{blue}{(8)}: permissible lower limit for the RAC policy ${o} = -0.06 e^{-0.03t} - 0.04$, \textcolor{blue}{(9)}: permissible upper limit for the RAC policy ${o} = 0.06 e^{-0.03t} + 0.04$, \textcolor{red}{(10)}: forbidden performance for the RAC control policy.}
\label{safe_erro_both}
\end{figure}

As demonstrated in Sections \ref{expdnn} and \ref{exppraac}, the trained DNN control policy exhibits better control accuracy compared to the stability-guaranteed RAC. In addition, as mentioned in Section \ref{aasafadfgadA}, in practice, heavy-duty WMRs with complex actuation mechanisms operate predominantly under nominal conditions, where the proposed LM-trained DNN control policy demonstrated high performance in Sections \ref{expdnn}. This control policy can tolerate minor or transient external disturbances without significant performance degradation. However, to ensure safe and reliable operation—especially in rare but critical scenarios involving severe disturbances or faults—it is essential to guarantee robustness and system stability in compliance with stringent international safety standards. Therefore, both control approaches were combined for the studied MPD to ensure not only accuracy but also robustness (see \textbf{Algorithm 4}). By integrating the stability-guaranteed RAC policy, the system benefits from two separate safety layers, maintaining continuous operation even when the trained DNN control policy fails to respond to potential disturbance attacks. This approach enables sustained functionality until an appropriate time for system overhaul. To evaluate practical performance, the proposed control provided in Section \ref{aasafadfgadA} were deployed on the experimental WMR platform to follow the S-shaped path shown in Fig. \ref{salam_s_shap}. The velocity reference for each wheel was generated using the method described in Ref. \cite{shahna2025lidar}. Figure \ref{safe_erro_both} illustrates the tracking error of the WMR under the combined control of the LM-trained DNN and the safe RAC policy. The plot presents the tracking errors for the left and right wheels, $e_L$ and $e_R$, over a duration of 180 s. Initially, the MPD operated under the DNN control policy, delivering strong performance while being monitored by performance bounds—indicated by dashed blue lines—defined as $\zeta = -0.02 e^{-0.35 t} - 0.02$ and $\zeta = 0.02 e^{-0.35 t} + 0.02$. At $t=82$ s, we introduced two side moderate signals as external disturbances to the operating system by increasing the control input by $15 \%$ relative to the demanded values. 

This caused the LM-trained DNN control tracking error to exceed the low-level safety layer bounds. Thus, the system transitioned at approximately $80.001$ s to the stability-guaranteed RAC policy, maintaining operation under the high-level safety layer. The RAC policy imposes its own bounds: ${o}_i=-0.06 e^{-0.04 t}-0.03$ (7) and ${o}_i=$ $0.06 e^{-0.03 t}+0.04(8)$, which are more relaxed to handle disturbances. This figure demonstrates the controller's ability to switch modes to preserve safety and adapt to disturbances, validating the effectiveness of the supervisory architecture.

\begin{figure}[h!]
\hspace*{-0.0cm} 
\centering
\scalebox{1.0}{\includegraphics[trim={0cm 0.0cm 0.0cm 0cm},clip,width=\columnwidth]{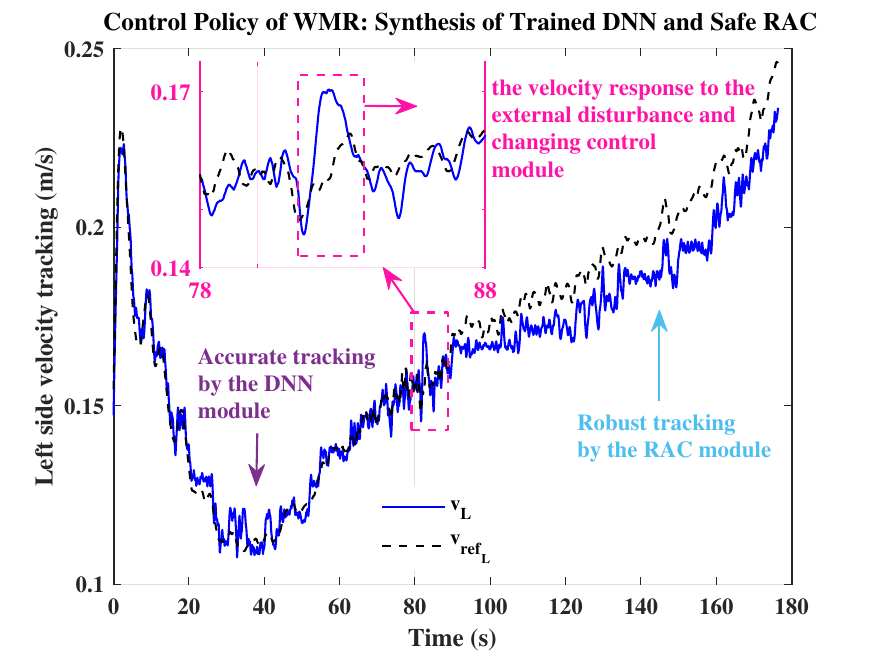}}
\caption{Exp. 3. Tracking on the left side under the synthesis of LM-trained DNN and safe RAC policy.}
\label{safe_vel_left}
\end{figure}

\begin{figure}[h!]
\hspace*{-0.0cm} 
\centering
\scalebox{1.0}{\includegraphics[trim={0cm 0.0cm 0.0cm 0cm},clip,width=\columnwidth]{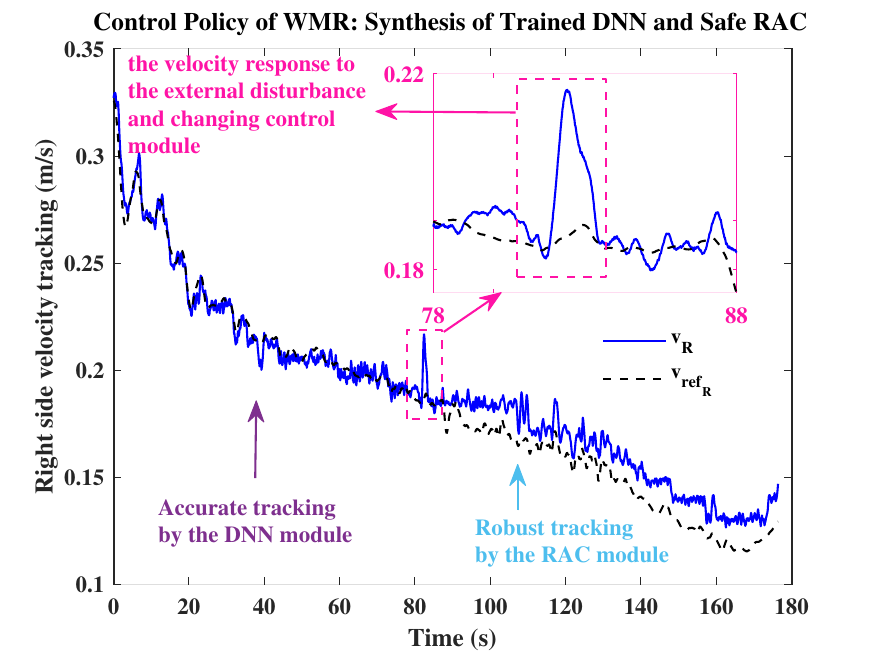}}
\caption{Exp. 3. Tracking in the right side under the synthesis of LM-trained DNN and safe RAC policy.}
\label{safe_right_vel}
\end{figure}

\begin{figure}[h!]
\hspace*{-0.0cm} 
\centering
\scalebox{1.0}{\includegraphics[trim={0cm 0.0cm 0.0cm 0cm},clip,width=\columnwidth]{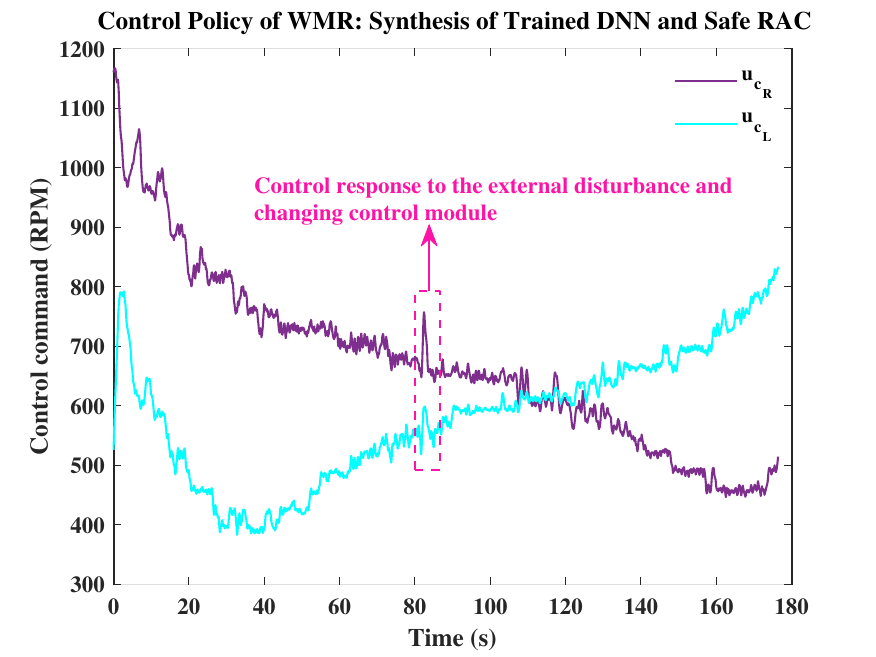}}
\caption{Exp. 3. Control commands (PMSM output) under the synthesis of LM-trained DNN and safe RAC policy.}
\label{con_eff_safe}
\end{figure}

Figure \ref{safe_vel_left} shows the left-side velocity tracking performance in this experiment. Initially, the DNN policy provided accurate tracking with minimal error, as annotated. At around 82 s, an external disturbance occured, causing a transient deviation. This triggered a latched functional switch from the DNN to the RAC policy, highlighted by the boxed area and velocity fluctuation between 78 and 88 s (with velocity varying between 0.14 and 0.17 $\mathrm{m} / \mathrm{s}$ ). After the transition, the RAC policy takes over and delivers robust tracking, adapting to the disturbance while maintaining alignment with the reference. 

The velocity tracking performance of both the right and left sides of the WMR control system-combining the DNN-based control policy and the safe RAC policy-demonstrates a consistent structure over time: accurate tracking by the DNN policy during the nominal phase (before $\sim82\mathrm{~s}$ ), a robust recovery phase managed by the RAC policy ( $\sim 82 \mathrm{~s}$ to $84 \mathrm{~s}$ ), and continued tracking by the RAC policy (after $84 \mathrm{~s}$), albeit at the cost of reduced accuracy. The right side generally exhibits higher velocity magnitudes, faster recovery, and smoother transitions post-disturbance, whereas the left side shows slightly more oscillation and slower stabilization. Overall, both plots validate the proposed control strategy's ability to manage the trade-off between system responsiveness and robustness under two safety layers with different authorities.

Figure \ref{con_eff_safe} shows the control commands for the left and right sides (PMSM output) of the WMR under the proposed control architecture. The signals $u_{c_L}$ (light blue) and $u_{c_R}$ (dark black) represent the rpm inputs to the motors over a duration of $180$ s. Initially, both control signals correspond to the DNN control policy $\left(u_{\mathrm{DNN}_i}\right)$, managing the system under nominal conditions until $t=82$ s. After this point, the control shifts to the RAC policy $\left(u_{s_i}\right)$, which compensates for the perturbed operation. This transition is marked by a small spike in the control inputs, indicating corrective action in response to the disturbance.

\section{Conclusion}
\label{section:conclusion}
This paper presents a practical and theoretically grounded solution for safe and robust control of heavy-duty WMRs operating under complex and unpredictable conditions, drawing inspiration from ISO/IEC TR 5469 - Artificial intelligence — Functional safety and AI systems.
To manage the fundamental trade-off between system robustness and system responsiveness, a hierarchical control policy was designed, and then monitored by two safety layers with differing levels of authority. An LM-trained DNN control policy served as the main controller under nominal conditions, delivering high-precision performance. When external disturbances occurred and became sufficiently intense such that the control performance no longer met the predefined low-level safety layer, a conditional latched function deactivated the LM-trained DNN control policy and activated a stability-guaranteed RAC strategy as an alternative, thereby ensuring continued safe operation. The high-level safety layer continuously monitored system performance during both control policies and triggered a shutdown only when the disturbances became so severe that compensation was no longer feasible, and continued operation would pose a risk to the system or its environment.
The proposed architecture ensured the uniform exponential stability of the entire framework of WMR. The control framework was further validated through step-by-step experimental evaluation of the standalone DNN-based policy, the standalone RAC policy, and their integration on a 6,000 kg WMR equipped with a five-stage actuation mechanism. Comparative results demonstrated the feasibility of all proposed approaches in real-world scenarios. This approach could pave the way for broader use of learning algorithms in heavy-duty robotic systems while still meeting stringent international standards.

\section*{Appendix A: LM Training For Dnn Control Policy}
\label{Matlab_DNN}
For data preparation in MATLAB, we define the input and target signals and splits the data into training, validation, and testing sets using \textit{dividerand}, as
\begin{lstlisting}[language=Matlab, caption={Raw data preparation and splitting}, label={lst:data_split}]
% Raw input/output data
X = v_i;   % 1xN inputs (linear velocity of each side of the WMR)
T = n_{p_i};   % 1xN targets (PMSM velocity as the control input RMP)
N = size(X, 2);

% Split data: for example: 70% train, 15% val, 15% test
[trainInd, valInd, testInd] = dividerand(N, 0.70, 0.15, 0.15);

% Extract subsets
Xtrain = X(:, trainInd); Ttrain = T(:, trainInd);
Xval   = X(:, valInd);   Tval   = T(:, valInd);
Xtest  = X(:, testInd);  Ttest  = T(:, testInd);
\end{lstlisting}

Then, inputs and targets are normalized using \textit{mapminmax} for better training performance. The parameters are computed only from the training set to avoid data leakage.

\begin{lstlisting}[language=Matlab, caption={Input and target normalization using mapminmax}, label={lst:normalization}]
% Normalize inputs to [-1, 1] based on training data
[XtrainN, inputPS] = mapminmax(Xtrain, -1, 1);
XvalN  = mapminmax('apply', Xval, inputPS);
XtestN = mapminmax('apply', Xtest, inputPS);

% Normalize targets similarly
[TtrainN, outputPS] = mapminmax(Ttrain, -1, 1);
TvalN  = mapminmax('apply', Tval, outputPS);
TtestN = mapminmax('apply', Ttest, outputPS);
\end{lstlisting}

As part of the network definition and data assignment a feedforward neural network with five hidden layers is created. Built-in preprocessing is disabled, and data indices are manually specified for each subset.

\begin{lstlisting}[language=Matlab, caption={Network creation and configuration}, label={lst:network_config}]
% Define network with 5 hidden layers
hiddenSizes = [30 25 15 10 5];
net = feedforwardnet(hiddenSizes, 'trainscg');

% Disable internal normalization
net.inputs{1}.processFcns = {};
net.outputs{end}.processFcns = {};

% Manual index assignment for training control
net.divideFcn = 'divideind';
net.divideParam.trainInd = 1 : numel(trainInd);
net.divideParam.valInd   = numel(trainInd) + (1:numel(valInd));
net.divideParam.testInd  = numel(trainInd) + numel(valInd) + (1:numel(testInd));

% Concatenate normalized data
XallN = [XtrainN, XvalN, XtestN];
TallN = [TtrainN, TvalN, TtestN];
\end{lstlisting}

Training parameters such as goal, minimum gradient, and maximum epochs are set. The network is then trained on the full normalized dataset.

\begin{lstlisting}[language=Matlab, caption={Training parameters and network training}, label={lst:train}]
% Set training options
net.trainParam.goal     = 1e-3;
net.trainParam.min_grad = 1e-4;
net.trainParam.epochs   = 200;

% Train the network
[net, tr] = train(net, XallN, TallN);
\end{lstlisting}

Finally, the trained network is evaluated on all data subsets. Predictions are denormalized and compared with the ground truth. Final MSE values are printed.

\begin{lstlisting}[language=Matlab, caption={Evaluate performance and compute MSE}, label={lst:evaluate}]
% Get actual index ranges
trainIdx = net.divideParam.trainInd;
valIdx   = net.divideParam.valInd;
testIdx  = net.divideParam.testInd;

% Predict (normalized)
YtrainN = net(XallN(:, trainIdx));
YvalN   = net(XallN(:, valIdx));
YtestN  = net(XallN(:, testIdx));

% Reverse normalization
Ytrain = mapminmax('reverse', YtrainN, outputPS);
Yval   = mapminmax('reverse', YvalN, outputPS);
Ytest  = mapminmax('reverse', YtestN, outputPS);

% Compute raw MSE
mse_train = mean((Ytrain - Ttrain).^2);
mse_val   = mean((Yval   - Tval).^2);
mse_test  = mean((Ytest  - Ttest).^2);

fprintf('MSE (raw scale): Train = %.3g, Val = %.3g, Test = %.3g\n', ...
        mse_train, mse_val, mse_test);
\end{lstlisting}

\section*{Appendix B: Blf-Based Safety Layer For DNN Policy}
\label{Matlab_BLF2}
As a logarithmic BLF serves as the safety monitor for the LM-trained DNN-based control policy of the MPD, we can implement it in MATLAB as follows:
\begin{lstlisting}[language=Matlab, caption={Safety index using logarithmic expression for DNN controllers}, label={lst:safety_log_check}]
% Define prescribed control performance threshold (zeta)
% Define input error signal e_i (can be scalar or vector)

% Compute the denominator of the safety expression
denom = (zeta)^2 - e_i.^2;

% Check if any value causes the expression to become undefined
if any(denom <= 0)
    error('Safety violation: e_i^2 >= (zeta)^2. Operation terminated.');
else
    safety_metric = log((zeta)^2 ./ denom);
    disp('safety_metric is defined. Operation can continue.');
    disp(safety_metric);
end
\end{lstlisting}

\section*{Appendix C: Implementing the Proposed RAC Policy}
\label{Matlab_RAC}
 MATLAB code for RAC is as follows:
\begin{lstlisting}[language=Matlab, caption={A sample form of MATLAB simulation for Eqs. (32) and (39)}]
% Define parameters of the RAC: delta_i, gamma_i, k_i
% as well as control performance metrics: o_shoot, o_bound, o_star
% Define e_i(t) as an input function of time
% Define time span for simulation
tspan = [t0 tf];  % Replace t0 and tf with start and end times

% Initial condition
theta0 = hat_theta_0;  % Initial value of \hat{\theta}_i(0)

% Define o(t) as:
o_func = @(t) (o_shoot - o_bound) * exp(-o_star * t) + o_bound;

% Define the ODE for \dot{\hat{\theta}}_i
dtheta_dt = @(t, theta) -delta_i * theta + ...
    gamma_i * (e_i / (o_func(t)^2 - e_i^2))^2;

% Solve the ODE
[t, theta] = ode45(dtheta_dt, tspan, theta0);

% Compute u_{s_i} = n_{p_i}
% u_{s_i} = n_{p_i} = -0.5 * k_i * e_i - gamma_i * (e_i / (o(t)^2 - e_i^2)) * theta
u_si = @(t, e_i, theta) -0.5 * k_i * e_i - gamma_i * (e_i / (o_func(t)^2 - e_i^2)) * theta;
\end{lstlisting}

\section*{Appendix D: Blf-based Safety Layer For Rac Policy}
\label{Matlab_BLF}

This logarithmic barrier function serves as the safety metric and can be implemented in MATLAB as follows:
\begin{lstlisting}[language=Matlab, caption={Safety index using logarithmic expression for RAC}, label={lst:safety_log_check}]
% Define prescribed control performance threshold (o)
% Define input error signal e_i (can be scalar or vector)

% Compute the denominator of the safety expression
denom = o^2 - e_i.^2;

% Check if any value causes the expression to become undefined
if any(denom <= 0)
    error('Safety violation: e_i^2 >= o^2. Operation terminated.');
else
    safety_metric = log(o^2 ./ denom);
    disp('safety_metric is defined. Operation can continue.');
    disp(safety_metric);
end
\end{lstlisting}

\section*{Appendix E: Proposed Synthesis of the DNN with Safe RAC Policy}
\label{Matlab_DNN_RAC}

If we define the low-level safety layer margin as $R = b^2 -e^2_i$, we can code $\alpha_1$ in MATLAB, as \eqref{52}.
\begin{lstlisting}[language=Matlab, caption={Low-level safety layer based on latch-off logic}, label={lst:safety_latch}]
% Initialize state: system starts active, deactivates permanently when R <= 0
alpha_1 = zeros(size(R));
active = true;

for i = 1:length(R)
    if active && R(i) <= 0
        active = false;  % Latch-off: deactivate forever once violated
    end
    % Allow operation only if still active and R > 0
     alpha_1(i)(i) = double(active && R(i) > 0);
end
\end{lstlisting}
This code creates a permanent shutdown signal that starts as "on" (active = true), and turns "off" permanently when the safety condition $\mathrm{R}<=0$ is violated for the first time. Even if the condition recovers later, the system remains off. In contrast, we also use $\alpha_2$ for the proposed RAC controllers that starts the latched-ON logic function at 0 and switches to 1 the moment the condition ($R \geq 0$) occurs, staying at 1 forever regardless of future inputs. $\alpha_2$ is defined as
\begin{equation}
\small
\begin{aligned}
\alpha_2(t)= \begin{cases}0, & \text { if } e^2_i(\tau)<b^2(\tau) \text { for all } \tau \in[0, t] \\ 1, & \text { if there exists } \tau \in[0, t] \text { such } b^2(\tau) \leq e^2(\tau)\end{cases}
\end{aligned}    
\end{equation}
We can code $\alpha_2$ in MATLAB, as

\begin{lstlisting}[language=Matlab, caption={Low-level safety layer based on latch-on logic}, label={lst:safety_latch_on}]

% Initialize latched state: start inactive, activate permanently when R <= 0
alpha_2 = zeros(size(R));
latched = false;

for i = 1:length(R)
    if ~latched && R(i) <= 0
        latched = true;  % Latch-on: activate permanently once violated
    end
    alpha_2(i) = double(latched);
end
\end{lstlisting}
; now we can define $u_{c_i}$:
\begin{lstlisting}[language=Matlab, caption={Computation of control input \( u_{c_i} \)}, label={lst:uc_control}]
% Define alpha1(t), alpha2(t), u_DNN_i(t), and u_s_i(t) as function handles

% Compute the control input u_{c_i} at time t
u_c_i = @(t) alpha1(t) * u_DNN_i(t) + alpha2(t) * u_s_i(t);
\end{lstlisting}

\bibliographystyle{IEEEtran}
\bibliography{MEHDI}

\begin{IEEEbiography}[{\includegraphics[width=1in,height=1.25in,clip,keepaspectratio]{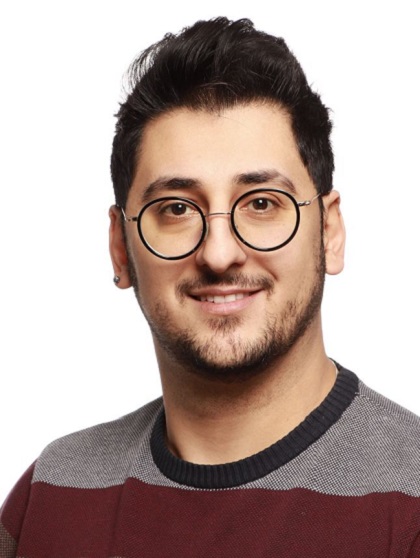}}]{Mehdi Heydari Shahna} earned a B.Sc. in electrical engineering from Razi University, Kermanshah, Iran, in 2015 and an M.Sc. in control engineering at Shahid Beheshti University, Tehran, Iran, in 2018. Since December 2022, he has been pursuing his doctoral degree in automation technology and mechanical engineering at Tampere University, Tampere, Finland. His research interests encompass robust control, robotics, fault-tolerant algorithms, and system stability, Robot Learning.
\end{IEEEbiography}

\begin{IEEEbiography}[{\includegraphics[width=1in,height=1.25in,clip,keepaspectratio]{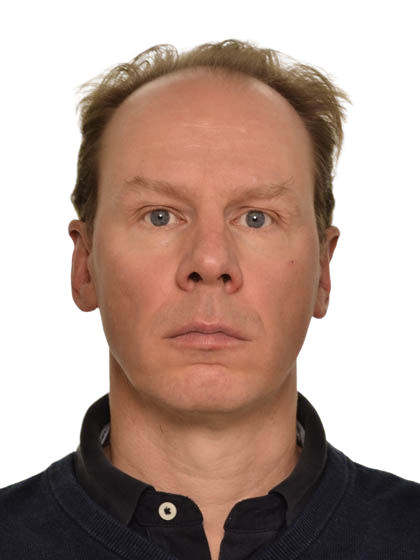}}]{Jouni Mattila} received an M.Sc. and Ph.D. in automation engineering from Tampere University of Technology, Tampere, Finland, in 1995 and 2000, respectively. He is currently a professor of machine automation with the Unit of Automation Technology and Mechanical Engineering at Tampere University. His research interests include machine automation, nonlinear-model-based control of robotic manipulators, and energy-efficient control of heavy-duty mobile manipulators.
\end{IEEEbiography}

\vfill

\end{document}